\documentclass{article} % For LaTeX2e
\usepackage{iclr2026_conference,times}

\usepackage{amsmath,amsfonts,bm}

\def\eqref#1{equation~\ref{#1}}
\def\1{\bm{1}}

\DeclareMathAlphabet{\mathsfit}{\encodingdefault}{\sfdefault}{m}{sl}
\SetMathAlphabet{\mathsfit}{bold}{\encodingdefault}{\sfdefault}{bx}{n}

\usepackage{hyperref}

\hypersetup{
  colorlinks=true,
  linkcolor=black,
  citecolor=black,
  urlcolor=black
}
\usepackage{url}
\usepackage{footmisc}
\usepackage{booktabs}
\usepackage{multirow} 
\usepackage{amsmath,amssymb}
\usepackage{xcolor}
\usepackage{float}
\usepackage[ruled,vlined]{algorithm2e}
\usepackage{colortbl}
\usepackage{graphicx}
\usepackage{tikz}
\usepackage{caption}
\usepackage{enumitem}
\usepackage{xspace}
\usepackage{subcaption}
\usepackage{xspace}
\usepackage{wrapfig}

\renewcommand{\refname}{References}  % article class
\renewcommand{\bibname}{References} 

\makeatother

\title{LLMBoost: Make Large Language Models Stronger with Boosting\\
}

\author{
\centering
Zehao Chen$^{1,2}$\thanks{Co-Leadership. \{zehaochenacid, chaojidouding\}@buaa.edu.cn; \{gyy\_chenzehao, aitianxiang, gyy\_liyifei\}@chinatelecom.cn} \thanks{Core Contributors. \{zehaochenacid, chaojidouding, lgxma01\}@buaa.edu.cn; \{gyy\_chenzehao, aitianxiang, gyy\_liyifei, gyy\_ligongxun\}@chinatelecom.cn},
Tianxiang Ai$^{2}$%
\footnotemark[1] \footnotemark[2],
Yifei Li$^{1,2}$%
\footnotemark[1] \footnotemark[2],,
Gongxun Li$^{1,2}$%
\footnotemark[2],
Yuyang Wei$^{2}$,
Wang Zhou$^{2}$,
Guanghui Li$^{2}$,
Bin Yu$^{2}$,
Zhijun Chen$^{1}$,
Hailong Sun$^{1}$,
Fuzhen Zhuang$^{1}$,
Jianxin Li$^{1}$\thanks{Corresponding author.  \{yikunb, dqwang, lijx\}@buaa.edu.cn },
Deqing Wang$^{1}$\footnotemark[\value{footnote}],
Yikun Ban$^{1}$\footnotemark[\value{footnote}]
\\
Beihang University$^1$\\
China Telecom eSurfing Cloud$^2$
}
\usepackage{amsthm}
\newtheorem{theorem}{Theorem}
\newtheorem{lemma}{Lemma}  
\newtheorem{corollary}{Corollary}

\newtheorem{definition}{Definition}

\newtheorem{assumption}{Assumption}
\usepackage{mathtools}  % 提供:= 
\usepackage{wrapfig}
\usepackage{algpseudocode}
\usepackage{algorithmicx}
\usepackage{makecell}
\SetKwFor{ForPar}{parfor}{do}{end}

\iclrfinalcopy % Uncomment for camera-ready version, but NOT for submission.
\begin{document}

\begingroup
\renewcommand\thefootnote{}
\footnotetext{This work was submitted to iclr 2026.}
\endgroup

\maketitle

\newcommand{\modelname}{\textsc{LLMBoost}\xspace}
\newcommand{\dianxindataset}{Chinatelecom Cloud Agent Dataset}
\newcommand{\dianxindatasetshort}{CCAD}
\newcommand{\dianxinlab}{a specific laboratory}

\begin{abstract}

Ensemble learning of LLMs has emerged as a promising alternative to enhance performance, but existing approaches typically treat models as black boxes, combining the inputs or final outputs while overlooking the rich internal representations and interactions across models.Ensemble learning of LLMs has emerged as a promising alternative to enhance performance, but existing approaches typically treat models as black boxes, combining the inputs or final outputs while overlooking the rich internal representations and interactions across models.
In this work, we introduce \modelname, a novel ensemble fine-tuning framework that breaks this barrier by explicitly leveraging intermediate states of LLMs. Inspired by the boosting paradigm, \modelname incorporates three key innovations. First, a \textit{cross-model attention mechanism} enables successor models to access and fuse hidden states from predecessors, facilitating hierarchical error correction and knowledge transfer. Second, a \textit{chain training paradigm} progressively fine-tunes connected models with an error-suppression objective, ensuring that each model rectifies the mispredictions of its predecessor with minimal additional computation. Third, a \textit{near-parallel inference paradigm} pipelines hidden states across models layer by layer, achieving inference efficiency approaching single-model decoding.
We further establish the theoretical foundations of \modelname, proving that sequential integration guarantees monotonic improvements under bounded correction assumptions. Extensive experiments on commonsense reasoning and arithmetic reasoning tasks demonstrate that \modelname consistently boosts accuracy while reducing inference latency.

\end{abstract}
\addtocontents{toc}{\protect\setcounter{tocdepth}{-1}}
\section{Introduction}

%\ban{Add citations for the following @zehao}
Large Language Models (LLMs) have revolutionized natural language processing, building on the Transformer’s attention mechanism Large Language Models (LLMs) have revolutionized natural language processing, building on the Transformer’s attention mechanism \citep{vaswani2017attention}. Individual models such as GPT-4 . Individual models such as GPT-4 \citep{achiam2023gpt}, Llama-3 , Llama-3 \citep{touvron2023llama}, and Qwen , and Qwen \citep{yang2025qwen3} have demonstrated remarkable performance across a wide range of domains. have demonstrated remarkable performance across a wide range of domains.
 Meanwhile, ensemble learning of LLMs — combining multiple agents to yield stronger, more robust performance — is gaining attention \citep{chen2025harnessing, lu2024merge}. However, despite its promise, the exploration of ensembles in the LLM era remains relatively nascent.

%Nevertheless, training LLMs requires massive computational resources that remain beyond the reach of most institutions. Instead of pushing toward ever-larger individual models, ensemble learning of pretrained LLMs \citep{chen2025harnessing,lu2024merge} has emerged as a promising alternative for building stronger systems, yet this direction remains underexplored.

Most existing LLM ensemble methods treat each model as a black box and fuse only their final outputs—either at the token level or the full-response level \citep{chen2025harnessing}—via majority voting \citep{li2024more,yu2024breaking}, weighted aggregation \citep{li2024purifying}, or routing strategies \citep{shnitzer2023llm}. However, these approaches ignore the rich information embedded in the hidden representations of the LLMs, and more fine-grained interactions—beyond mere output fusion—are needed to fully exploit complementary knowledge across models.

In this paper, we go beyond conventional output-level ensembling and propose a novel fine-tuning framework for LLM ensembles, \modelname, which enables fine-grained internal interactions by progressively leveraging the hidden representations within each model in the boosting style. Our main contributions are summarized as follows.

%Inspired by the “boosting” paradigm, \modelname extracts and leverages the internal states of LLMs during ensemble learning, progressively fine-tuning each model to complement the under-learned knowledge of its predecessors.

[\textbf{Model Architecture}]. We design a residual cross-model attention mechanism in which successor models can adaptively access and fuse the intermediate hidden states of predecessor models, rather than relying solely on its final outputs. The cross-model attention is constructed at the layer level, breaking the black-box barrier and enabling state-driven hierarchical interaction among LLMs. In addition, progressively targeted fusion is applied to output logits, which allows \modelname to transfer valid knowledge across models effectively.

[\textbf{Chain Training}]. We introduce a sequential training framework for connected LLMs,
%, where a backbone model is trained with LoRA adapters to preserve general knowledge from pretraining, 
and each subsequent model is fine-tuned based on the training log of its predecessor. Inspired by the boosting paradigm, the role of a successor model is to rectify the key tokens of its predecessor. Thus, we propose an error-suppression objective, which suppresses the maximal (incorrect) logit produced by the predecessor while enhancing the correct one. By doing so, \modelname significantly reduces the overall computational requirement while ensuring that performance is progressively boosted across the chain of models.

[\textbf{Near-Parallel Inference}].
We propose a near-parallel streaming inference paradigm to minimize the latency introduced by sequential interactions among LLMs. Instead of waiting for the predecessor to complete full sequence generation, \modelname transmits hidden states at the layer level without delay: 
the hidden state from the $(l-1)$-th layer of the predecessor is immediately passed to the $l$-th layer of the successor for parallel computation, even before the forward pass is completed. 
This pipelined design significantly reduces inference latency, achieving performance close to that of a single model while substantially outperforming vanilla inference ensembles.

[\textbf{Theoretical Analysis}].
We provide a formal theoretical analysis for \modelname, establishing a principled foundation for its sequential error-correction mechanism. We prove that, under the condition that each new model effectively corrects its predecessor's errors, the ensemble's performance is guaranteed to improve with each model added. This guarantee holds when the contribution of each new model is appropriately weighted, demonstrating a clear theoretical path toward progressively enhancing prediction accuracy through our boosting-style integration. Our analysis is grounded in Mean Squared Error, offering a direct and comprehensive measure of performance gains.

[\textbf{Empirical Performance}] .
Extensive experiments on commonsense and arithmetic reasoning tasks — including a custom set co-developed with \dianxinlab — confirm our approach’s effectiveness. Compared to other ensemble baselines, \modelname consistently enhances performance across tasks in tests: it delivers an overall average improvement of 3.9\%, with an even larger 6.6\% as the maximum improvement, with 3\% improvement observed specifically on \dianxindataset(\dianxindatasetshort). Moreover, our near-parallel decoding strategy reduces inference latency by 47\% compared to conventional sequential ensembles, demonstrating clear advantages in efficiency.

\section{Related Work}

\textbf{LLM Ensemble Methods}.
Existing LLM ensemble methods primarily focus on output fusion using various strategies.
First, inference-time methods enable dynamic integration during decoding, such as token-level probability aggregation \citep{yao2024determine,yu2024breaking,xu2024bridging,li2024purifying} and span-level integration \citep{xu2024hit,fu2025rlae}. Other methods operate on the final outputs through majority voting \citep{li2024more}, weighted fusion \citep{guha2024smoothie}, routing-based selection \citep{ong2024routellm,sikeridis2025pickllm}, and prompt-based ensembles \citep{he2024llm}.
Second, post-inference methods fuse complete outputs, either via non-cascaded aggregation \citep{li2024more,guha2024smoothie} or cascaded strategies such as difficulty-aware revision \citep{yue2023large}.
Third, pre-inference methods focus on task–model matching, including pretrained routing \citep{ong2024routellm,shnitzer2023llm} and non-pretrained routing \citep{sikeridis2025pickllm,zhang2025avengers}.
While effective, these approaches treat models largely as independent black boxes, overlooking the internal representations and cross-model interactions that could further enhance ensemble performance.

\textbf{Boosting Methods.}
Boosting \citep{friedman2001greedy,hastie2009elements,ke2017lightgbm} is a classical ensemble paradigm that builds strong learners by sequentially correcting the errors of prior learners. Gradient Boosting minimizes loss by fitting new learners to the residuals of previous outputs \citep{hastie2009elements,chen2016xgboost}, while AdaBoost reweights samples to emphasize misclassified instances and combines learners via weighted voting \citep{freund1997decision,li2008adaboost}. Recent work has extended boosting to LLMs: \cite{agrawal2025ensemw2s} applied AdaBoost to train a series of weak LLMs into a stronger model\footnote{We did not find open-source code, and thus we exclude it from experimental comparisons.}, but this approach ensembles over training samples while still ignoring internal interactions among LLMs. \cite{zou2025transformer} introduced copilot adapters to correct errors during model fine-tuning, but the scope is limited to a single auxiliary adapter.

Distinct from existing LLM ensemble works, our work advances the field by ensembling on internal representations. \modelname enables subsequent models to access and fuse predecessors’ intermediate hidden states—not just final outputs—for precise error correction, addressing the gap of insufficient intermediate interaction in existing methods while ensuring stable performance gains.

%Internal-model methods remain relatively underexplored, with few leveraging intermediate hidden states for fine-grained interaction. Our work advances internal-model approaches with a token-level dynamic correction framework rooted in boosting; it enables subsequent models to access and fuse predecessors’ intermediate hidden states—not just final outputs—for precise error correction, addressing the gap of insufficient intermediate interaction in existing methods while ensuring stable performance gains.

%Gradient Boosting applies gradient descent to minimize loss, with each new learner fitting residuals of previous ensemble outputs \cite{friedman2001greedy}\cite{friedman2002stochastic}\cite{hastie2009elements}. XGBoost optimizes this with regularization and parallel processing \cite{chen2016xgboost}, while LightGBM boosts efficiency via histogram-based splitting and GPU acceleration \cite{ke2017lightgbm}, collectively enhancing predictive performance in supervised learning.

%\ban{To do: add EnsembleW2S, add T-Copilot}

\section{Proposed Framework: \modelname}
\label{method}

\begin{figure*}[!t]
% \captionsetup{font=footnotesize}
  \centering
  % \vspace{-0.05in}
  \includegraphics[width= 1.0 \textwidth]{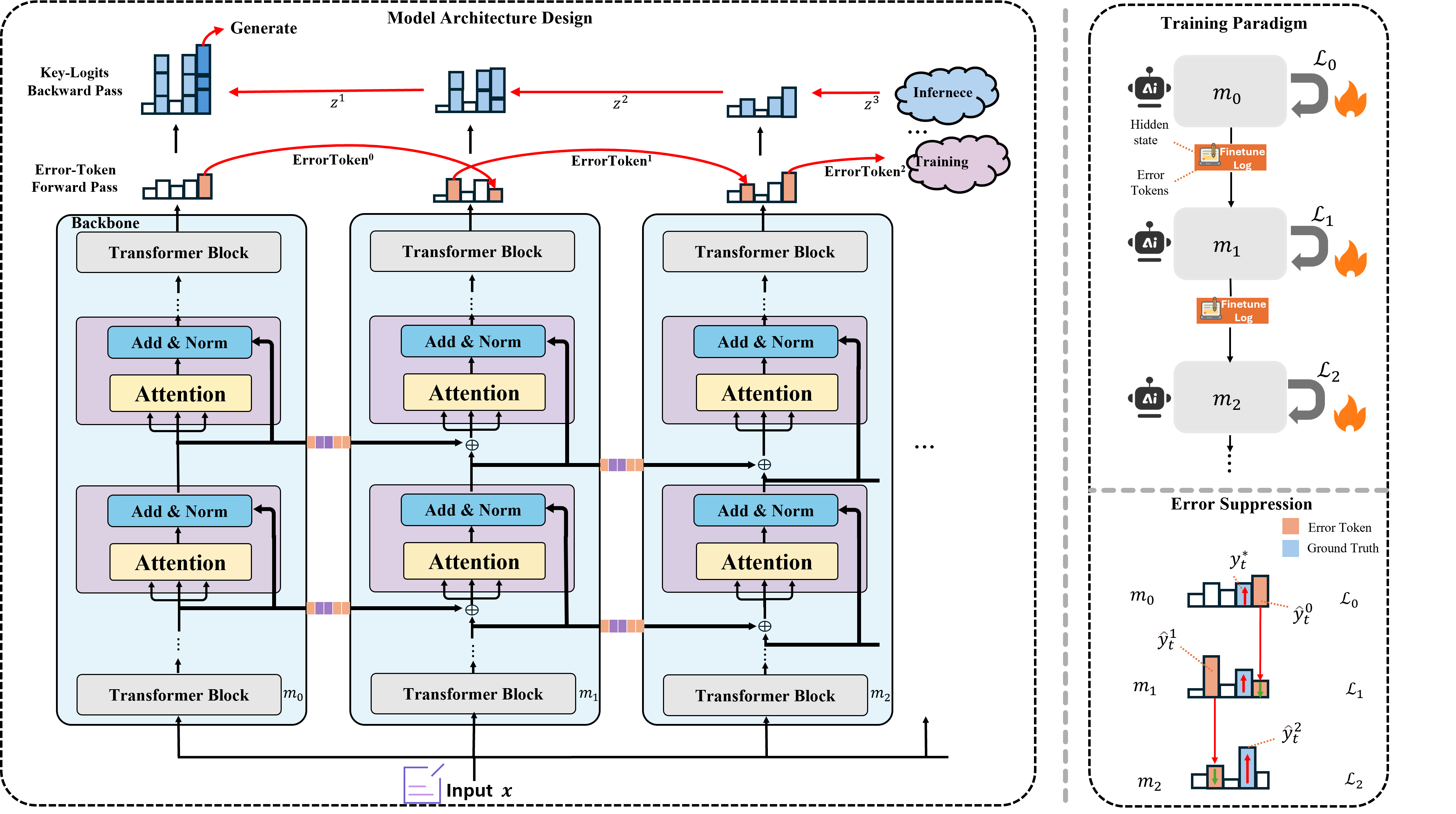}
  \caption{\modelname Framework. The overall framework comprises three key components: (1) \modelname Model Design, (2) Training Paradigm, and (3) Error-Suppression Objective.}
  \label{fig:overall_framework}
  \vspace{-10pt}
\end{figure*}

We introduce \modelname, a hierarchical ensemble fine-tuning framework that enables LLMs to perform internal interactions during both training and inference. In the following sections, we elaborate on \modelname from three perspectives: its architecture, training paradigm, and inference paradigm.

%Our approach is inspired by the core idea of XGBoost \cite{chen2016xgboost}—progressive refinement—but breaks away from its classical formulation through a series of systematic modifications. Specifically, we transform the traditional residual correction into \textbf{fine-grained token-level refinement}, and optimize the original serial dependency into a \textbf{hierarchy-aware pipelined parallelism}. In addition, we design an \textbf{error-aware loss function} that, while preserving the standard cross-entropy supervision, imposes extra constraints on tokens mispredicted by the previous model, thereby enabling more targeted optimization. This design not only enhances robustness but also improves inference efficiency. \\

\textbf{Notations}. As illustrated in Figure~\ref{fig:overall_framework}, \modelname is designed as a sequential ensemble of pretrained LLMs, where each LLM adopts a decoder-only Transformer architecture. Suppose there are $(1+n)$ LLMs in total. Formally, the ensemble can be represented as
$\mathcal{M} = \{ m_0(\cdot; \theta_0), m_1(\cdot; \theta_1), \dots, m_n(\cdot; \theta_n) \}$,
where we use $m_i \in \mathcal{M}$ to denote the $i$-th model and $\theta_i$ its trainable parameters for simplicity.
In one fine-tuning round $\tau$, we sample an input–output pair $(X_\tau, Y_\tau)$ from a data distribution $D_{\mathcal{X}, \mathcal{Y}}$, where $X_\tau$ is the input sequence and $Y_\tau$ the corresponding target sequence. \modelname generates the output sequence $\widehat{Y}_\tau$ in an autoregressive manner, aiming to approximate the target sequence $Y_\tau$. 
At the $t$-th decoding step, let $z^{(i)}_t$ denote the logits produced by the $i$-th model, and let $p^{(i)}_t = \mathrm{softmax}(z^{(i)}_t)$ be the corresponding probability distribution over the vocabulary. 
The predicted token of the $i$-th model is then
$\hat{y}^{(i)}_t = \arg\max p^{(i)}_t,$
while the ground-truth token is denoted by $y^{\ast}_t$.

\subsection{Framework Architecture}
%Our proposed \textbf{\modelname} framework is built upon decoder-only Transformer architectures, with model parameters initialized from publicly available HuggingFace checkpoints.

%In the following sections, we will elaborate on the model design, the training paradigm, and the inference paradigm, respectively.

% \iffalse

% \begin{wrapfigure}{r}{0.45\textwidth} % r=右侧，宽度自调
%   \centering
%   \vspace{1em}
%   \includegraphics[width=0.43\textwidth]{iclr2026/pictures/modeldesign& trainingParadigm.png}
%   \caption{\modelname Framework. The overall framework comprises two key components: \textbf{(1) \modelname Model Design}, \textbf{(2) Training Paradigm}, and \textbf{ Inference Paradigm} (details are provided in Fig.~\ref{fig:inference}).}
%   \label{fig:overall_framework}
% \end{wrapfigure}
% \fi
%Each sub-model in \modelname is initialized from public pre-trained checkpoints, while lightweight modules are newly introduced to support task adaptation and cross-model interaction. 
%Formally, the ensemble can be represented as:
%$\mathcal{M} = \{ m_0(\cdot; \theta_0), m_1(\cdot; \theta_1), \dots, m_K(\cdot; \theta_k) \}$,
%where $m_i$ denotes the $i$-th model and $\theta_i$ its trainable parameters. For simplicity, we use $m_i$ to denote $m_i(\cdot;\theta_i)$ in the following.

%\subsubsection{Fine-grained Token-level Refinement}
\paragraph{Cross-Model Attention.}
A key innovation of \modelname lies in its fine-grained token-level refinement mechanism.
Unlike conventional ensembles where models operate independently, each model $m_i$ directly accesses and fuses the hidden representations of its predecessor $m_{i-1}$, thereby enabling hierarchical interaction across the $n$ LLMs.
Concretely, at the $t$-th decoding step, let $h^{(i-1)}_{l-1,t}$ denote the hidden state from the $(l-1)$-th Transformer layer of the $(i-1)$-th model. The cross-model attention mechanism at the $l$-th layer of the $i$-th model is defined as:
\begin{equation}
\label{equ:hiddenStatesAddNorm}
    \tilde{h}^{(i)}_{l,t} = 
\begin{cases}
\mathrm{LayerNorm}\!\big(h^{(i)}_{l-1,t} + h^{(i-1)}_{l-1,t}\big), & \text{if} \ l \bmod \eta = 0, \\[6pt]
h^{(i)}_{l-1,t}, & \text{otherwise}.
\end{cases}
\end{equation}
where $\eta$ is an integer constant that controls the sparsity of cross-model connections among the $n$ LLMs.
With the enhanced hidden states, the cross-attention output is calculated by:

\begin{equation}
   \hat{h}^{(i)}_{l,t} 
= \mathrm{softmax}\!\Big(\tfrac{Q^{(i)}_{l,t} (K^{(i)}_{l,t})^\top}{\sqrt{d_k}}\Big) V^{(i)}_{l,t}, \ \ \text{where} \ Q^{(i)}_{l,t} = \tilde{h}^{(i)}_{l,t} W_Q, 
\quad K^{(i)}_{l,t} = \tilde{h}^{(i)}_{l,t} W_K, 
\quad V^{(i)}_{l,t} = \tilde{h}^{(i)}_{l,t} W_V,
\end{equation}
followed by another residual enhancement:
\begin{equation}
h^{(i)}_{l,t} = \mathrm{LayerNorm}\!\big( \hat{h}^{(i)}_{l,t}+ \tilde{h}^{(i)}_{l,t} \big).
\end{equation}

This design equips later models with structured access to their predecessors’ reasoning process, yielding:
 \emph{error correction}, where inaccurate states are adjusted, and 
 \emph{knowledge reinforcement}, where correct signals are amplified.  

For the input and output streaming, the input sequence $X_\tau$ is shared across all models in \modelname, while the output fusion streaming incorporates two key innovations.

(1) \textbf{Error-Token Forward Pass}. 
To enable successor models to recognize and correct token-level errors made by their predecessors, we introduce an error-token forwarding mechanism. This design improves both performance and robustness of \modelname. During training, the error logits produced by model $m_i$ are sequentially passed to its successor $m_{i+1}$ under a specific learning objective (detailed in the next section).
At the $t$-th decoding step, let $\hat{y}^{(i)}_t = \arg\max p^{(i)}_t$ denote the predicted token index, where the error tokens are defined as follows:
\begin{equation}
\text{ErrorToken}^{(i)}_t =
\begin{cases}
\hat{y}^{(i)}_t, & \text{if } \hat{y}^{(i)}_t \neq y_t^{\ast}, \\[6pt]
\emptyset, & \text{otherwise},
\end{cases}
\end{equation}
where $y_t^{\ast}$ denotes the ground-truth token. Thus, only incorrect token indices are passed forward, encouraging successors to explicitly learn from predecessor mistakes.

(2) \textbf{Key-Logits Backward Pass}.
To further stabilize the inference process, we propose a top-$k$ logits fusion strategy, where later successor models pass key logits backward to the base model $m_0$.  
Instead of aggregating over the entire vocabulary, each sub-model contributes only its most informative token logits, thereby reducing the influence of noisy, low-confidence outputs.  
Formally, given the logit distribution $z_t^{(i)}$ from the $i$-th model $m_i$, we define the restricted candidate set and the ensemble update rule as:
\begin{equation} 
\label{equ:logitsFusion}
\begin{aligned} 
z_t &= z^{(0)}_t + \sum_{i=1}^{n} \lambda_i \cdot \mathrm{Mask}(z^{(i)}_t, \mathcal{V}_{\text{top-}k}^{(i)}), \ \ \text{where} \ \mathcal{V}_{\text{top-}k}^{(i)} = \mathrm{\arg \text{top-k}}(z^{(i)}_t),  
\end{aligned} 
\end{equation} 
where $z_t$ denotes the final raw logits, $\lambda_i$ is a scaling coefficient, and is a scaling coefficient, and $\mathrm{Mask}(\cdot)$ restricts contributions to the top- restricts contributions to the top-$k$ candidates by setting all other indices to zero in each logit distribution. candidates by setting all other indices to zero in each logit distribution.

\subsection{Chain Training Paradigm}
\textbf{Error-Suppression Objective.}
To enable each model to refine its predecessor by correcting erroneous predictions while preserving correct ones, we introduce a token-level suppression loss.  
The principle is that each $m_i$ should preserve plausible predictions from $m_{i-1}$, while intervening 
only when necessary. Formally,  we define the error-suppression loss:
\begin{equation}
\label{equ:preference_optimization}
    \mathcal{L}_{s,t} = - \log\!\Big( \sigma\!\big( \beta \,[ \log p^{(i)}_t[y_t^{ \ast}] - \log p_t^{(i)}[\text{ErrorToken}^{(i-1)}_t]] \big) \Big) \cdot \mathbf{1}\!\{ \text{ErrorToken}^{(i-1)}_t\neq \emptyset \},
\end{equation}
where $\sigma$ is the sigmoid, $\beta$ a scaling factor, and the indicator $\mathbf{1}$ activates the loss only at positions where the predecessor $i-1$ predicts incorrectly.
The overall objective for $m_i$ becomes:
\begin{equation}
\label{equ:totalloss}
\mathcal{L}_i = \sum_t   \mathcal{L}_{s,t}    - \alpha\sum_t \log p^{(i)}_{t}[y_t^{ \ast}],
\end{equation}
where the right item is the cross-entropy loss, and $\alpha$ is a balance coefficient.  
This dual-objective strategy endows the ensemble with a \emph{correction-while-retaining} ability, 
preserving useful signals while refining erroneous ones.

\begin{algorithm}[htbp]
\caption{\modelname: Training Process}
\label{alg:chained_ensemble}
\SetAlgoLined
\LinesNumbered
% \KwIn{Base model list $\mathcal{M} = \{m_0(\cdot; \theta_0), m_1(\cdot; \theta_1), \dots, m_n(\cdot; \theta_n)\}$, training dataset $\mathcal{D}$, Log pool $g$}
% \KwOut{Fine-tuned model list $\mathcal{\tilde{M}}= \{\tilde{m}_0(\cdot; \theta_0), \tilde{m}_1(\cdot; \theta_1), \dots, \tilde{m}_n(\cdot; \theta_n)\}$}
\KwIn{Base model list $\mathcal{M} = \{m_0, m_1, \dots, m_n\}$, training dataset $\mathcal{D}$, }
\KwOut{Fine-tuned model list $\mathcal{\tilde{M}}= \{\tilde{m}_0, \tilde{m}_1, \dots, \tilde{m}_n\}$}
% \textbf{Stage 1: Train the base model $m_0(\cdot; \theta_0)$}\;
% Initialize $m_0(\cdot; \theta_0) \gets$ ApplyLoRA($m_0(\cdot; \theta_0)$, $\phi$)\;
Initialize log pool $g$ \;
\textbf{Stage 1: Train the base model $m_0$}\;
\For{each batch $(x, y^{\ast}) \in \mathcal{D}$}{
    % Forward pass: $z^{(0)} \gets m_0(x; \theta_0)$\;
    Forward pass: $z^{(0)} \gets m_0(x)$\;
    Compute Cross-Entropy loss:$\mathcal{L}_0 \gets \text{CrossEntropyLoss}(z_0, y^{\ast})$\;
    Update $m_0$ parameters via LoRA using $\mathcal{L}_0$\;
    % Fine-tune $m_0$ via LoRA using $\mathcal{L}_0$: $\tilde{m}_0 \gets m_0$\;
    % Update $m_0(\cdot; \theta_0)$ parameters via LoRA fine-tuning using $\mathcal{L}_0$\;
    % Record training log: $g \gets \text{UpdateLog}(g, m_0,\{h^{(0)}, \text{ErrorToken}^{(0)}\})$
}
Obtain fine-tuned model: $\tilde{m}_0 \gets m_0$\;
\textbf{Stage 2: Train subsequent models $\{m_1(\cdot; \theta_1), \dots, m_n(\cdot; \theta_n)\}$}\;
\For{$i = 1$ \KwTo $n$}{
    % Initialize $m_i(\cdot; \theta_i) \gets$ ApplyLoRA($m_i(\cdot; \theta_i)$, $\phi$)\;
    \For{each batch $(x, y^{\ast}) \in \mathcal{D}$}{
        Get previous model's inference trajectory: $\{h^{(i-1)},\text{ErrorToken}^{(i-1)}\} \gets \tilde{m}_{i-1}(x)$\;
        % Get hidden states $h^{(i-1)}$ and predicted tokens $z^{(i-1)}$ from $\tilde{m}_{i-1}(x)$\;
        % Get $m_{i-1}$ training log:  $\{h^{(i-1)}, \text{ErrorToken}^{(i-1)}\} \gets \text{ReadLog}(g, m_{i-1})$\;
        Forward pass: $z^{(i)} \gets m_i(x,h^{(i-1)})$\;
        Compute Error-Suppression loss $\mathcal{L}_i$ via \ref{equ:totalloss} using $\{ z^{(i)},y^{\ast},\text{ErrorToken}^{(i-1)}\}$\;
        Update $m_i$ parameters via LoRA using $\mathcal{L}_i$\;
        % Fine-tune $m_i$ via LoRA using $\mathcal{L}_i$: $\tilde{m}_i \gets m_i$\;
        % Update $m_i(\cdot; \theta_0)$ parameters via LoRA fine-tuning using $\mathcal{L}_i$\;
        % Record training log: $g \gets \text{UpdateLog}(g, m_i,\{h^{(i)}, \text{ErrorToken}^{(i)}\})$
    }
    Obtain fine-tuned model: $\tilde{m}_i \gets m_i$\;
}
\KwRet{$\{\tilde{m}_0(\cdot; \theta_0), \tilde{m}_1(\cdot; \theta_1), \dots, \tilde{m}_n(\cdot; \theta_n)\}$}
\end{algorithm}

\textbf{Training Pipeline}.
% To enable efficient adaptation of large ensembles, we adopt a \emph{two-stage, chained fine-tuning} strategy (Algorithm~\ref{alg:chained_ensemble}). This design ensures that each sub-model learns not only from ground-truth supervision but also from the reasoning trajectory of its predecessor.
Our model adopts a two-stage chained fine-tuning strategy (see Algorithm~\ref{alg:chained_ensemble}), which ensures that each sub-model learns not only from ground-truth supervision but also from the reasoning trajectory of its predecessor.
In the first stage, the base model $m_0$ is adapted to the target task following standard  is adapted to the target task following standard LoRA fine-tuning. The parameter updates of  fine-tuning. The parameter updates of $m_0$ are solely driven by the input data  are solely driven by the input data $x$ and the ground-truth labels  and the ground-truth labels $y^{\ast}$.
In the second stage, subsequent models $m_i $ are fine-tuned in a chained manner. Specifically, for each model, we first extract the internal hidden states $h^{(i-1)}$ and error tokens $\text{ErrorToken}^{(i-1)}$ from its predecessor $\tilde{m}_{i-1}$. The input $x$, together with $h^{(i-1)}$, is then fed into $m_i$ for training. The Error-Suppression loss is computed using $z^{(i)}$,  $y^{\ast}$, and $\text{ErrorToken}^{(i-1)}$. This mechanism enables later models to progressively correct the reasoning errors of earlier models, thereby improving the overall ensemble performance.

\begin{figure*}[!th]
% \captionsetup{font=footnotesize}
  \centering
  \includegraphics[width= 0.8\textwidth]{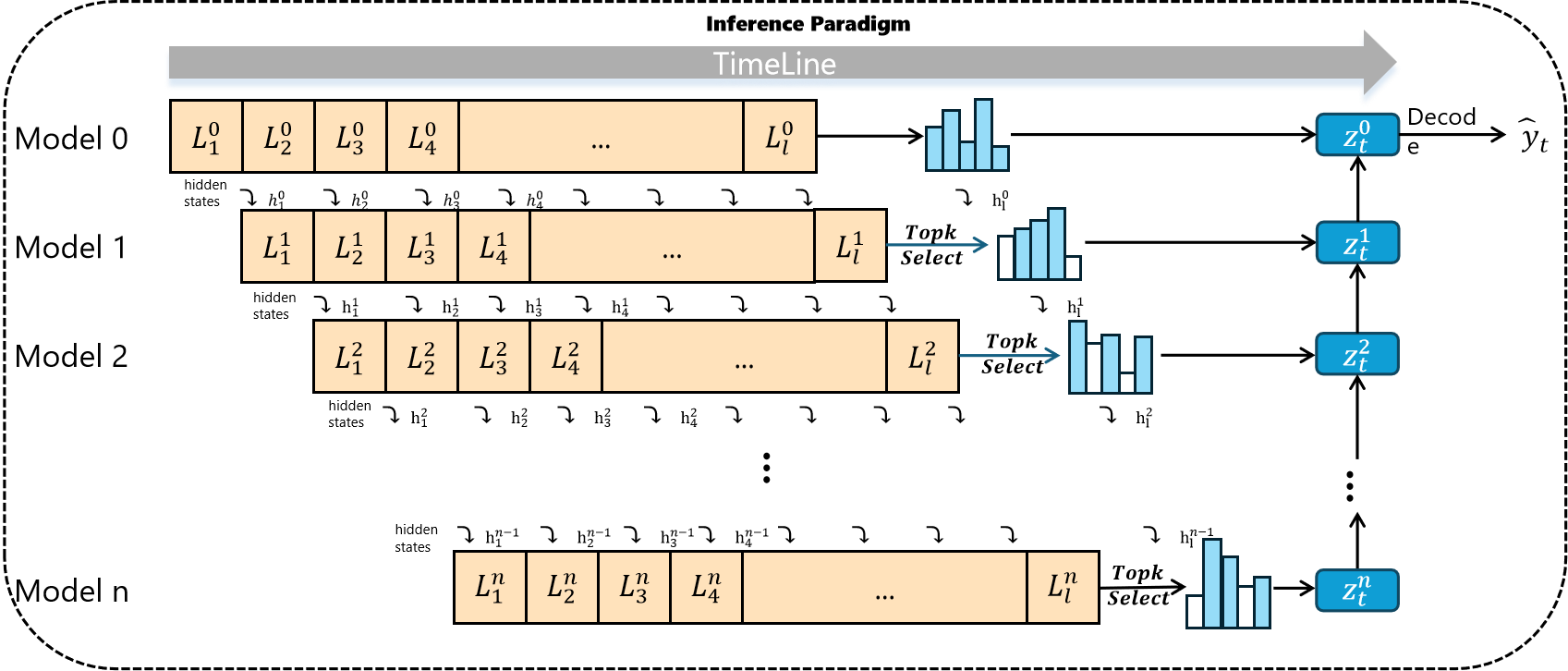}
  \caption{\textbf{Near-parallel inference paradigm.} 
The figure uses the timeline as the axis to show the computation state of each model at each moment. Hidden states are propagated across layers and models, and logits are aggregated and decoded into the final prediction, enabling overlapped computation and reduced inference latency.}
  \label{fig:inference}
  \vspace{-10pt}
\end{figure*}

\subsection{Near-Parallel Inference Paradigm} 

As each successor model $m_i$ depends on the intermediate hidden states of its predecessor $m_{i-1}$, a naive decoding scheme requires $m_i$ to begin its decoding only after $m_{i-1}$ has completed the entire decoding process. This results in an inference time complexity of $O(n)$ if one model takes $O(1)$ time per sequence, which becomes impractical when $n$ is large.

To address this, we propose a near-parallel inference
% layer-wise parallel streaming
paradigm (Algorithm~\ref{alg:parallel_inference_simple} in Appendix \ref{app:Near_Parallel_Algorithm}), which enables efficient inference across multiple models. Instead of waiting for the predecessor to finish full sequence generation, \modelname streams hidden states at the \emph{layer level}: the hidden state from the $(l-1)$-th layer of $m_{i-1}$ is immediately passed to the $l$-th layer of $m_i$ while the forward pass of $m_{i-1}$ is still ongoing (see Figure~\ref{fig:inference}).
For implementation, we maintain a thread-safe shared hidden state pool $g$ for inter-model communication. At each decoding step $t$ and each layer $l$, model $m_i$ waits for the corresponding hidden state $h^{(i-1)}_{l-1,t}$ from its upstream model $m_{i-1}$. Once $h^{(i-1)}_{l-1,t}$ becomes available, it is pushed into $g$ and immediately consumed by $m_i$ to compute $h^{(i)}_{l,t}$.

Finally, after all $(n+1)$ models complete their forward passes at step $t$, their logits are fused following the Key-Logits Backward Pass rule. The final logits $z_t$ are computed according to Eq.~\ref{equ:logitsFusion}, from which the next token is generated.

\section{Theoretical Guarantees: Why Sequential Correction Works}
\label{sec:analysis}

In the previous section, we introduced \modelname, a framework built on the principle of sequential error correction. A natural and critical question arises: does this boosting-style integration of models guarantee a performance improvement? This section provides a formal theoretical justification covering two key aspects: (1) Inference Efficacy, demonstrating that adding a new model provably reduces prediction error under a validity condition; and (2) Optimization Stability, proving that our dual-objective training guarantees a descent path on the primary task.

Our analysis focuses on the Mean Squared Error (MSE) between the ensemble's predicted probability distribution and the ground truth. We encapsulate the requirement for effective correction in a core assumption. For brevity, we present the main results here; full proofs and detailed derivations are provided in Appendix~\ref{app:proof_of_thm_and_cor}.

\textbf{Notations for Analysis}. At decoding step $t$, let $A_i$ denote the sampling space of $m_i$.
We consider an ensemble $\mathcal{M}_i$ with accumulated logits $\hat{z}^{(i)}_t$, yielding predicted distribution $\hat{P}^{(i)}_t=\mathrm{softmax}(\hat{z}^{(i)}_t)$.
Let $\hat{y}^{(i-1)}_{<t}$ be the sequence generated by $\mathcal{M}_{i-1}$ up to step $t-1$, and let $p^*_t(\cdot \mid X_\tau,\hat{y}^{(i-1)}_{<t})$ denote the ground-truth distribution, represented by the one-hot vector $p^\ast_t$.
The residual error of $\mathcal{M}_{i-1}$ is defined as $e^{(i-1)}_t=p^\ast_t - \hat{P}^{(i-1)}_t$.
  To quantify how a new model $m_i$ influences the ensemble's prediction, we define its effective linear contribution to the probability space as $g^{(i)}_t \triangleq J_{\mathrm{softmax}}(\hat{z}^{(i-1)}_t) z^{(i)}_t$, where $J_{\mathrm{softmax}}(\hat{z}^{(i-1)}_t)$ is the Jacobian matrix of the softmax function evaluated at $\hat{z}^{(i-1)}_t$. 

\subsection{Inference Efficacy: Error Reduction Guarantee}
% inference-version
\begin{assumption}
\label{ass:correction_efficacy}
For any model $i \geq 1$, any decoding step $t$, and any vocabulary dimension $v \in \{1, \dots, |\mathcal{V}|\}$, we assume that the successor model $m_i$ produces an effective linear contribution $g^{(i)}_t$ that is a valid corrective term of the residual error vector $e^{(i-1)}_t$. Specifically, we assume its systematic bias in fitting this residual is less than the total uncertainty of the residual. Formally, given a generated context $(X_\tau, \hat{y}^{(i-1)}_{<t})$:
\begin{equation}
\epsilon_{g^{(i)}_{t, v}} < \sqrt{\epsilon_{i-1, t, v}^2 + \sigma_{i-1, t, v}^2},
\end{equation}
where we let $\Theta_{i-1} := \{\theta_0, \dots, \theta_{i-1}\}$ be the parameters of the ensemble $\mathcal{M}_{i-1}$, drawn from a joint distribution $\mathcal{A}_{i-1}$. The bias ($\epsilon^2$) and variance ($\sigma^2$) terms are defined as:

\begin{align}
% Bias of the preceding ensemble M_{i-1} at inference
\epsilon_{i-1, t, v}^2 &:= \mathbb{E}_{(X_\tau,Y_\tau)\sim\mathcal{D}} \left[ \left( p^*_{t,v} - \mathbb{E}_{\Theta_{i-1} \sim \mathcal{A}_{i-1}}[\hat{P}^{(i-1)}_{t,v} \mid X_\tau, \hat{y}^{(i-1)}_{<t}] \right)^2 \right] < \infty, \\
% Variance of the preceding ensemble M_{i-1} at inference
\sigma_{i-1, t, v}^2 &:= \mathbb{E}_{(X_\tau,Y_\tau)\sim\mathcal{D}} \left[ \mathrm{Var}_{\Theta_{i-1} \sim \mathcal{A}_{i-1}}[\hat{P}^{(i-1)}_{t,v} \mid X_\tau, \hat{y}^{(i-1)}_{<t}] \right] < \infty , \\
% Bias of the new model m_i's contribution h^i_t in fitting the residual at inference
\epsilon_{g^{(i)}_{t, v}}^2 &:= \mathbb{E}_{\substack{\Theta_{i-1} \sim \mathcal{A}_{i-1} \\ (X_\tau,Y_\tau)\sim\mathcal{D}}} \left[ \left( e^{(i-1)}_{t,v} - \mathbb{E}_{\theta_i \sim A_i}[g^{(i)}_{t,v} \mid \Theta_{i-1}, X_\tau, \hat{y}^{(i-1)}_{<t}] \right)^2 \right] < \infty.
\end{align}
Furthermore, we assume that the logits generated by the successor model have a finite fourth moment, i.e., $\mathbb{E}[\|z^{(i)}_t(\cdot \mid X_\tau, \hat{y}^{(i-1)}_{<t})\|_2^4] < \infty$.
\end{assumption}

This assumption posits that each boosting step is productive—the new model is more accurate at predicting the error than the existing ensemble is at predicting the target. Our chain training paradigm is explicitly designed to encourage this condition. Based on this assumption, we establish the following theorem and corollary.

\begin{theorem}
\label{thm:mse_reduction}
Under Assumption \ref{ass:correction_efficacy}, For each model $i \geq 1$, decoding step $t$, and vocabulary dimension $v$, there exists an upper bound $\lambda_{i, t, v}^* > 0$ for the scaling coefficient. For any weight $\lambda_{i} \in (0, \lambda_{i, t, v}^*)$, the expected squared error of the new ensemble $\mathcal{M}_{i}$ at inference is strictly lower than that of the predecessor ensemble $\mathcal{M}_{i-1}$ for that dimension. Specifically:
\begin{equation}
\begin{aligned}
\mathbb{E}_{\substack{\Theta_{i-1} \sim \mathcal{A}_{i-1} \\ \theta_i \sim A_i \\ (X_\tau,Y_\tau)\sim\mathcal{D}}}
\left[ \left( p^*_{t,v} - \hat{P}^{(i)}_{t,v} \right)^2 \mid X_\tau, \hat{y}^{(i-1)}_{<t} \right]
<
\mathbb{E}_{\substack{\Theta_{i-1} \sim \mathcal{A}_{i-1} \\ (X_\tau,Y_\tau)\sim\mathcal{D}}}
\left[ \left( p^*_{t,v} - \hat{P}^{(i-1)}_{t,v} \right)^2 \mid X_\tau, \hat{y}^{(i-1)}_{<t} \right].
\end{aligned}
\end{equation}
\end{theorem}

\subsection{Optimization Stability: Guaranteed Descent}
\label{sec:optimization}

While Theorem~\ref{thm:mse_reduction} guarantees that a valid corrector improves inference performance, it relies on the premise that the model can be effectively trained to act as such a corrector. A potential concern with our dual-objective training (Eq.~\ref{equ:totalloss}) is that the error-suppression loss $\mathcal{L}_{s}$ might conflict with the primary cross-entropy loss $\mathcal{L}_{ce}$, potentially destabilizing the training.
To address this, we analyze the optimization landscape and show that a constructive descent path always exists.

We introduce a geometric assumption regarding the alignment between the primary gradient $g_{ce}$ and the error-suppression gradient $g_{s}$.

\begin{assumption}[Gradient Relative Alignment]
\label{ass:grad_align_short}
We assume the error-suppression gradient does not strictly oppose the primary gradient. Formally, there exist constants $0< \rho_i < 1$ and $\Gamma_i > 0$ such that $\langle g_{s}, g_{ce} \rangle \ge -\rho_i \|g_{s}\| \|g_{ce}\|$ and $\|g_{s}\| \le \Gamma_i \|g_{ce}\|$.
\end{assumption}

This assumption is mild in practice due to the mutual exclusivity of the softmax function (suppressing an error token naturally reallocates probability mass to the ground truth). Based on this, we establish the existence of a valid descent path:

\begin{theorem}[Guaranteed Descent on Primary Loss]
\label{thm:descent_short}
Under Assumption~\ref{ass:grad_align_short} and standard smoothness conditions, there exists a lower bound for the balancing coefficient $\alpha > \rho_i \Gamma_i$ and a corresponding learning rate range $(0, \eta^*(\alpha)]$ such that every gradient update step strictly decreases the primary cross-entropy loss $\mathcal{L}_{ce}$.
\end{theorem}

The detailed proof and constructive derivation of $\eta$ can be found in Appendix~\ref{app:support}.

\vspace{0.5em}
\noindent\textbf{Remark and Interpretation}.
Theorems \ref{thm:mse_reduction} and \ref{thm:descent_short} jointly constitute the theoretical foundation of \modelname. 
First, Theorem \ref{thm:mse_reduction} establishes the efficacy of the ensemble: it guarantees that as long as the new model acts as a valid corrector (bias $<$ uncertainty), the system's accuracy improves. The existence of the bound $\lambda$ reflects the classic bias-variance trade-off, where the ensemble weight must balance the correction signal against introduced variance.
Second, Theorem \ref{thm:descent_short} establishes the learnability of the model: it proves that despite the auxiliary error-suppression objective, a valid optimization path exists where the primary task performance monotonically improves.
Together, these results confirm that \modelname is not only effective in principle but also trainable in practice, setting the stage for the empirical validation in Section~\ref{sec:experiments}.

\section{Experiments}
\label{sec:experiments}
We conduct extensive experiments to evaluate the effectiveness and efficiency of our proposed method. 
First, we evaluate \modelname's performance on two representative tasks, Commonsense Reasoning and Arithmetic Reasoning. 
Second, to verify the generality of our approach, we further evaluate \modelname in an industrial scenario provided by the \textit{\dianxinlab}. 
Third, we evaluate the time efficiency of our near-parallel inference paradigm.
Additional experimental results and ablation studies are included in Appendix \ref{appendix: additionalexperiments}. \\
\textbf{Commonsense \& Arithmetic Datasets.} To comprehensively evaluate \modelname, we utilize a broad suite of reasoning tasks: (i) Commonsense reasoning: PIQA \citep{bisk2020piqa}, HellaSwag \citep{zellers2019hellaswag}, WinoGrande \citep{sakaguchi2021winogrande},BoolQ \citep{clark2019boolq}, SIQA \citep{sap2019socialiqa}, and OpenbookQA (OBQA) \citep{mihaylov2018can}.
(ii) Arithmetic reasoning: AQuA \citep{ling2017program},GSM8K \citep{cobbe2021training}, MAWPS \citep{koncel2016mawps}, and SVAMP \citep{patel2021nlp}. More details can be found in Appendix~\ref{appendix:datasets}.  \\
\textbf{\dianxindataset(\dianxindatasetshort).} We construct a dataset from real-world public cloud scenarios in \dianxinlab, targeting the evaluation of AI agents in toolchain scheduling. It assesses whether LLMs can interpret user intentions, plan multi-step toolchains, and execute tool calls under realistic industrial constraints. More details can be found in Appendix~\ref{appendix:datasets}.  \\ 
% Unlike existing benchmarks, our dataset introduces multi-tool dependencies, diverse task categories, and varying complexities, offering a more practical test of robustness, generalizability, and utility in tool orchestration.\\
\textbf{Baselines.} To clearly validate the effectiveness of the proposed method, experiments compare it with three mainstream large model optimization and ensembling approaches, namely VOTE \citep{li2024more}, UNITE \citep{yao2024determine}, and T-copilot \citep{zou2025transformer}. Detailed description of these three baseline methods can be found in Appendix \ref{app:baselines}. \\
% \textbf{Implementation Details.} For \modelname, we employ models from both the LLaMA-3 family (LLaMA-3.2-3B, LLaMA-3.1-8B) and the Qwen-2.5 family (Qwen-2.5-3B, Qwen-2.5-7B) as sub-models.  We further extend the generation function from the Huggingface Transformers library to support distributed, token-level inference acceleration across multiple sub-models. All experiments are conducted on four NVIDIA H800 GPUs. More details can be found in Appendix~\ref{app: implementationDetails}. 
%We adopt representative scales from both families as baselines for \textsc{\modelname} and evaluate the proposed framework on \textbf{LLaMA-3.2-3B, LLaMA-3.1-8B, Qwen-2.5-3B, and Qwen-2.5-7B}.
%, ensuring architectural consistency across ensembles by adopting identical decoder-only structures.
% During training, the AdamW optimizer and a linear learning rate scheduler are utilized.

% \input{iclr2026/tables/Matched-ScaleEnsembles}
\begin{table*}[!t]
    \centering
    \tiny
    \renewcommand{\arraystretch}{0.9}
    \caption{Performance comparison on \textbf{Arithmetic Reasoning} tasks . The proposed \modelname demonstrates robust improvements on AQuA, GSM8K, MAWPS, and SVAMP.}
    \label{tab:arithmetic_result}
    \small
    \resizebox{0.65\textwidth}{5.5cm}{
    \begin{tabular}{@{}l|c|c c c c c@{}}
        \toprule
        \multirow{2}{*}{Model} & \multirow{2}{*}{Params} & \multicolumn{5}{c}{Arithmetic Reasoning (Acc. $\uparrow$)} \\
        \cmidrule(l){3-7}
        & & AQuA & GSM8K & MAWPS & SVAMP & Avg. \\
        \midrule[0.35pt]
        
        % === Llama-3.1-8B ===
        \textbf{Llama-3.1-8B} & $1\times8B$ & 42.3 & 63.7 & 89.5 & 77.4 & 68.2  \\
        \cmidrule(lr){1-7}
        T-copilot & $2\times8B$ & 42.8 & 66.5 & 89.7 &  75.7 & 68.7 \\
        vote & $2\times8B$ & 44.4  & 67.8 & 89.5 & 77.5 & 69.8 \\
        UNITE & $2\times8B$ & 42.8 & 65.5 & 89.3 & 76.5 & 68.5 \\
        \rowcolor{gray!18}
        \modelname & $2\times8B$ & 46.8 & 68.8 & 90.1 & 80.1 & \underline{71.5} \\
        \cmidrule(lr){1-7}
        vote & $3\times8B$ & 43.9 & 66.0 &  90.7 & 78.1 & 69.7 \\
        UNITE & $3\times8B$ & 43.3 & 65.6 & 88.3 & 77.2 & 68.6 \\
        \rowcolor{gray!14} 
        \modelname & $3\times8B$ & 47.6 & 68.5 & 92.0 & 81.3 & \textbf{72.4} \\

        \midrule
        \midrule
        
        % === Llama-3.2-3B ===
        \textbf{LLama-3.2-3B} & $1\times3B$ & 31.9 & 49.7 & 84.9 & 65.8 & 58.1 \\
        \cmidrule(lr){1-7}
        T-copilot & $2\times3B$ & 33.3 & 49.5 & 84.6 & 66.7 & 58.5\\
        vote & $2\times3B$ & 34.6 & 50.9 & 85.4 & 68.2 & 59.8 \\
        UNITE & $2\times3B$ & 31.1 & 50.9 & 85.2 & 65.7 & 58.2 \\
        \rowcolor{gray!18}
        \modelname & $2\times3B$ & 35.8 & 51.5 & 87.4 & 70.1 & \underline{61.2} \\
        \cmidrule(lr){1-7}
        vote & $3\times3B$ & 32.9 & 53.5 & 87.4 & 66.4 & 60.1 \\
        UNITE & $3\times3B$ & 32.5 & 52.9 & 86.0 & 66.0 & 59.4 \\
        \rowcolor{gray!14} 
        \modelname & $3\times3B$ & 35.1 & 54.3 & 86.2 & 69.9 & \textbf{61.4} \\

        \midrule
        \midrule
        
        % === Qwen-2.5-7B ===
        \textbf{Qwen-2.5-7B} & $1\times7B$ & 56.8 & 75.1 & 92.0 & 86.1 & 77.5 \\
        \cmidrule(lr){1-7}
        T-copilot & $2\times7B$ & 56.9 & 75.6 & 92.7 & 85.3 & 77.6 \\
        vote & $2\times7B$ & 57.8 & 76.3 & 91.6 & 86.4 & 78.0 \\
        UNITE & $2\times7B$ & 59.5 & 76.3 & 92.2 & 86.4 & 78.6\\
        \rowcolor{gray!18}
        \modelname & $2\times7B$ & 59.8 & 78.1 & 94.1 & 88.8 & \underline{80.2} \\
        \cmidrule(lr){1-7}
        vote & $3\times7B$ & 60.9 & 77.4 & 92.4 & 86.7 & 79.4 \\
        UNITE & $3\times7B$ & 60.9 & 77.8 & 92.8 & 86.3 & 79.5 \\
        \rowcolor{gray!18}
        \modelname & $3\times7B$ & 61.8 & 78.9 & 93.3 & 88.4 & \textbf{80.6} \\

        \midrule
        \midrule
        
        % === Qwen-2.5-3B ===
        \textbf{Qwen-2.5-3B} & $1\times3B$ & 54.1 & 69.5 & 89.9 & 83.4 & 74.2 \\
        \cmidrule(lr){1-7}
        T-copilot & $2\times3B$ & 53.9 & 72.1 & 90.1 & 82.7 & 74.7 \\
        vote & $2\times3B$ & 54.2 & 71.2 & 89.5 & 82.6 & 74.4  \\
        UNITE & $2\times3B$ & 58.5 & 71.6 & 91.3 & 84.7 & 76.5 \\
        \rowcolor{gray!18}
        \modelname & $2\times3B$ & 60.2 & 74.0 & 92.4 & 85.3 & \underline{78.0} \\
        \cmidrule(lr){1-7}
        vote & $3\times3B$ & 59.7 & 71.9 & 92.1 & 85.2 & 77.2 \\
        UNITE & $3\times3B$ & 59.3 & 71.6 & 92.3 & 86.0 & 77.3 \\
        \rowcolor{gray!18}
        \modelname & $3\times3B$ & 61.2 & 73.4 & 92.6 & 85.6 & \textbf{78.2} \\
        
        \bottomrule
    \end{tabular}
    }
\end{table*}
\begin{table*}[!t]
    \centering
    \scriptsize
    \renewcommand{\arraystretch}{0.9}
    \caption{Performance comparison on \textbf{Commonsense Reasoning} tasks . \modelname consistently achieves higher accuracy across PIQA, WinoGrande, HellaSwag, BoolQ, SIQA, and OBQA compared to baselines.}
    \label{tab:commonsense_result}
    \small
    \resizebox{0.75\textwidth}{5.5cm}{
    \begin{tabular}{@{}l|c|c c c c c c c@{}}
        \toprule
        \multirow{2}{*}{Model} & \multirow{2}{*}{Params} & \multicolumn{7}{c}{Commonsense Reasoning (Acc. $\uparrow$)} \\
        \cmidrule(l){3-9} 
        & & PIQA & WinoG. & HellaS. & BoolQ & SIQA & OBQA & Avg. \\
        \midrule[0.35pt]
        
        % === Llama-3.1-8B ===
        \textbf{Llama-3.1-8B} & $1\times8B$ & 83.3 & 81.8 & 90.6 & 69.5 & 75.8 & 79.0 & 80.0 \\
        \cmidrule(lr){1-9}
        T-copilot & $2\times8B$ & 84.2 & 83.2 & 91.4 & 69.8 & 76.9 & 78.4 & 81.0 \\
        vote & $2\times8B$ & 81.7 & 81.3 & 86.8 & 67.3 & 76.1 & 84.8 & 79.7 \\
        UNITE & $2\times8B$ & 86.9 & 84.9 & 93.5 & 70.5 & 79.2 & 84.8 & 83.3 \\
        \rowcolor{gray!18}
        \modelname & $2\times8B$  & 87.9 & 86.1 & 94.5 & 71.7 & 81.6 & 85.6 & \underline{84.4} \\
        \cmidrule(lr){1-9}
        vote & $3\times8B$ & 87.7 & 83.2 & 90.3 & 71.8 & 80.4 & 85.4 & 83.1 \\
        UNITE & $3\times8B$ & 85.3 & 85.0 & 92.2& 70.9 & 79.9 & 85.0 & 83.1 \\
        \rowcolor{gray!14} 
        \modelname & $3\times8B$  & 87.4 & 86.3 & 95.2 & 72.5 & 80.7 & 84.8 & \textbf{84.5} \\

        \midrule
        \midrule
        
        % === Llama-3.2-3B ===
        \textbf{LLama-3.2-3B} & $1\times3B$ & 78.5 & 77.7 & 83.9 & 62.4 & 74.3 & 76.8 & 75.6 \\
        \cmidrule(lr){1-9}
        T-copilot & $2\times3B$ & 75.7 & 75.9 & 83.4 & 64.6 & 73.8 & 77.2 & 75.1 \\
        vote & $2\times3B$ & 79.4 & 76.9 & 84.7 & 66.2 & 75.4 & 75.8 & 76.4 \\
        UNITE & $2\times3B$ & 82.8 & 80.0 & 88.7 & 66.5 & 76.7 & 77.4 & 78.7 \\
        \rowcolor{gray!18}
        \modelname & $2\times3B$  & 84.7 & 81.4 & 90.4 & 67.1 & 78.0 & 78.8 & \underline{80.1} \\
        \cmidrule(lr){1-9}
        vote & $3\times3B$ & 80.5 & 76.4 & 85.1 & 66.7 & 77.5 & 77.2 & 77.2 \\
        UNITE & $3\times3B$ & 82.8 & 81.3 & 89.9 & 67.1 & 78.4 & 78.2 & 79.6 \\
        \rowcolor{gray!14} 
        \modelname & $3\times3B$  & 84.2 & 82.0 & 91.5 & 70.2 & 79.0 & 79.6 & \textbf{81.1} \\

        \midrule
        \midrule

        % === Qwen-2.5-7B ===
        \textbf{Qwen-2.5-7B} & $1\times7B$ & 86.7 & 83.9 & 89.4 & 68.7 & 77.6 & 87.4 & 82.3 \\
        \cmidrule(lr){1-9}
        T-copilot & $2\times7B$ & 87.1 & 84.3 & 90.2 & 70.8 & 79.3 & 89.2 & 83.5 \\
        vote & $2\times7B$ & 86.3 & 84.5 & 91.3 & 70.7 & 80.0 & 88.8 & 83.6 \\
        UNITE & $2\times7B$ & 86.1 & 84.1 & 91.6 & 68.6 & 79.1 & 88.2 & 83.0 \\
        \rowcolor{gray!18}
        \modelname & $2\times7B$ & 89.5 & 86.7 & 95.1 & 71.6 & 79.1 & 91.6 & \underline{85.6} \\
        \cmidrule(lr){1-9}
        vote & $3\times7B$ & 87.5 & 82.2 & 91.3 & 70.2 & 79.6 & 85.8 & 82.7 \\
        UNITE & $3\times7B$ & 86.1 & 83.3 & 91.5 & 70.3 & 78.4 & 86.8 & 82.7 \\
        \rowcolor{gray!18}
        \modelname & $3\times7B$ & 89.9 & 88.8 & 95.4 & 74.4 & 81.2 & 91.4 & \textbf{86.9} \\

        \midrule
        \midrule
        
        % === Qwen-2.5-3B ===
        \textbf{Qwen-2.5-3B} & $1\times3B$ & 84.5 & 80.1 & 80.6 & 68.1 & 77.7 & 84.6 & 79.3 \\
        \cmidrule(lr){1-9}
        T-copilot & $2\times3B$ & 83.9 & 80.2 & 82.6 & 68.9 & 78.4 & 86.4 & 80.1 \\
        vote & $2\times3B$ & 83.4 & 80.0 & 86.3 & 68.7 & 77.9 & 84.0 & 80.1 \\
        UNITE & $2\times3B$ & 84.1 & 80.6 & 83.5 & 67.3 & 78.0 & 83.0 & 79.4 \\
        \rowcolor{gray!18}
        \modelname & $2\times3B$ & 85.9 & 81.8 & 91.4 & 69.7 & 79.0 & 86.2 & \underline{82.3} \\
        \cmidrule(lr){1-9}
        vote & $3\times3B$ & 84.1& 79.7 & 87.6 & 67.8 & 77.4 & 83.8 & 80.4 \\
        UNITE & $3\times3B$ & 84.8 & 80.8 & 85.1 & 66.2 & 77.3 & 86.0 & 80.0 \\
        \rowcolor{gray!18}
        \modelname & $3\times3B$  & 86.1 & 82.0 & 91.7 & 70.4 & 80.0 & 87.6 & \textbf{83.0} \\
        
        \bottomrule
    \end{tabular}
    }
\end{table*}
\begin{table*}[!t]
    \centering
    \caption{Performance comparison (\%)  across both matched-scale and weakly-heterogeneous ensembles.}
    \label{tab:main_result}
    \small
    \resizebox{\textwidth}{!}{
    \begin{tabular}{@{}l l c c c c c c c l c c c c c l@{}}
        \toprule
        \multirow{2}{*}{Model} & \multirow{2}{*}{Params}& Latency & \multicolumn{7}{c}{Commonsense Reasoning (Acc. $\uparrow$)} & \multicolumn{5}{c}{Arithmetic Reasoning (Acc. $\uparrow$)} \\
        \cmidrule(l){4-10} \cmidrule(l){11-15}
        & & (s/token $\downarrow$) & PIQA & WinoG. & HellaS. & BoolQ & SIQA & OBQA & Avg. & AQuA & GSM8K & MAWPS & SVAMP & Avg. \\
        \midrule[0.35pt]
        Qwen2.5-7B & 7B & 0.056 & 86.7 & 83.9 & 89.4 & 68.7 & 77.6 & 87.4 & 82.3 & 56.8 & 75.1 & 92.0 & 86.1 & 77.5 \\
        \rowcolor{gray!14} \modelname(3B $\times$ 2) & 6B \textcolor{red}{\small(-1B)} & \textbf{0.043} & 85.9 & 81.8 & 91.7 & 70.4 & 80.0 & 86.2 & \textbf{82.3} & 60.2 & 74.0 & 92.4 & 85.3 & \textbf{78.0} \\
        \midrule
        Qwen2.5-14B & 14B & 0.098 & 90.2 & 85.8 & 92.9& 71.2 & 78.9 & 91.4 & 85.1 & 62.2 & 78.5 & 92.4 & 87.4 & 80.1\\
        \rowcolor{gray!14}  \modelname(7B $\times$ 2) & 14B & \textbf{0.075} & 89.5 & 86.7 & 95.1 & 71.6 & 79.1 & 91.6 & \textbf{85.6} & 59.8 & 78.1 & 94.1 & 88.8 & \textbf{80.2} \\

        \midrule[0.35pt]  \midrule[0.35pt]
        \multicolumn{14}{@{}c}{\textbf{\textit{Weakly-Heterogeneous Ensemble}}}\\
        \midrule
        LLama-3.2-3B & 3B & 0.027 & 78.5 & 77.7 & 83.9 & 62.4 & 74.3 & 76.8 & 75.6 & 31.9 & 49.7 & 84.9 & 65.8 & 58.1 \\
        Llama-3.1-8B & 8B & 0.038 &83.3 & 81.8 & 90.6 & 69.5 & 75.8 & 79.0 & 80.0 & 42.3 & 63.7 & 89.5 & 77.4 & 68.2  \\
        \rowcolor{gray!14} \modelname(8B + 3B) & 11B & 0.049 & 85.9 & 85.0 & 93.8 & 71.3 & 80.4 & 84.0 & \textbf{83.7} & 43.7 & 64.4 & 89.7 & 78.6 & \textbf{69.1}  \\
        \bottomrule
        
    \end{tabular}
    }
\vspace{-15pt}
\end{table*}

\subsection{Main Experiments Results}

\textbf{Performance on arithmetic reasoning datasets.}
Table~\ref{tab:arithmetic_result} summarizes the performance on arithmetic reasoning benchmarksunder LoRA training settings. 
\textsc{\modelname} shows a consistent advantage over competing methods, outperforming voting ensembles by a margin of 3.9\% and surpassing stronger baselines by significant margins. 
The effectiveness of our approach is further evidenced by the scaling analysis, where the ensemble achieves a 6.2\% increase in accuracy compared to the single-model baseline. 
Overall, \textsc{\modelname} proves to be a robust solution for enhancing mathematical reasoning capabilities across different model scales.

\textbf{Performance on commonsense datasets.}
Table~\ref{tab:commonsense_result} summarizes the results on commonsense reasoning tasks under LoRA training settings.
\modelname consistently outperforms both majority voting and the UNITE baseline across all six datasets. 
Most notably, it achieves substantial improvements on WinoGrande, surpassing the best baseline by 6.6\%. 
On average, \modelname improves commonsense reasoning accuracy by 1.7\% over the strongest baseline and 5.9\% over voting methods. 
The performance gains are consistent across different model scales, demonstrating that our sequential boosting framework reliably enhances the commonsense reasoning capabilities of base LLMs.

\begin{figure*}[t]
  \centering
  \begin{subfigure}{0.48\textwidth}
    \centering
    \includegraphics[width=\linewidth]{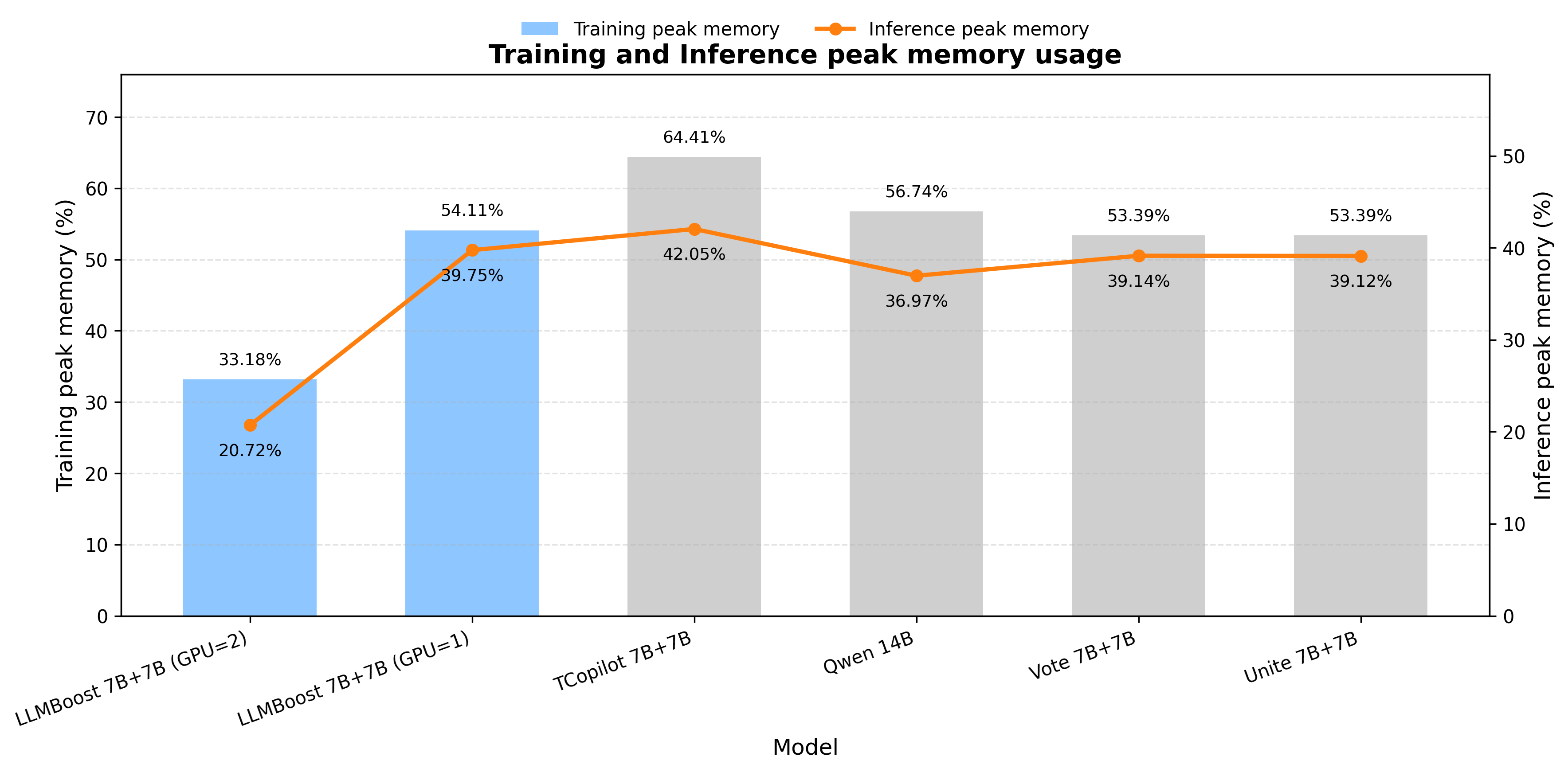}
    \caption{Training and inference peak memory usage across different models.}
    \label{fig:memory_usage}
  \end{subfigure}
  \hfill
  \begin{subfigure}{0.48\textwidth}
    \centering
    \includegraphics[width=\linewidth]{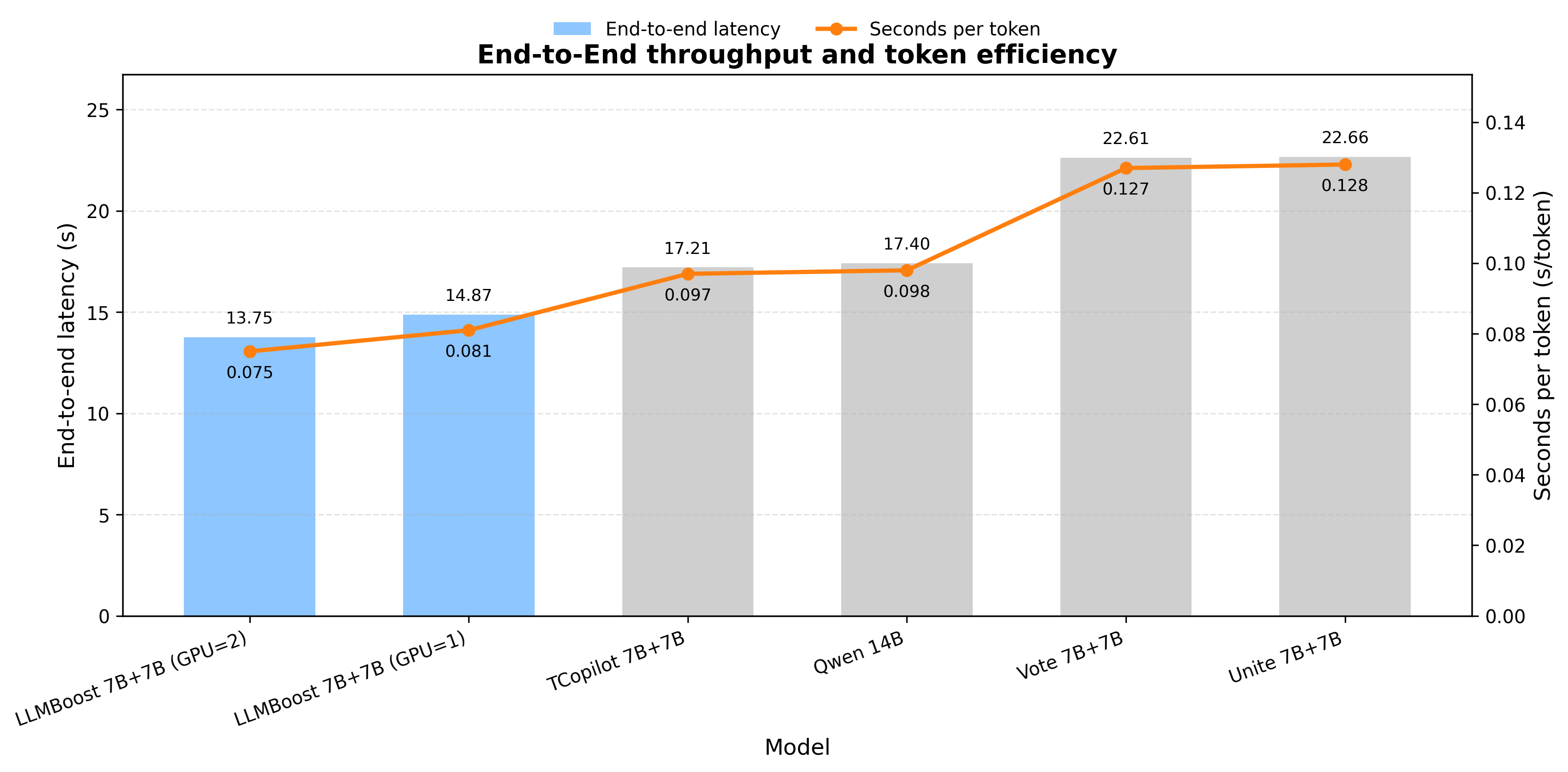}
    \caption{End-to-end latency and per-token generation cost.}
    \label{fig:latency_efficiency}
  \end{subfigure}
  \caption{Overall system efficiency comparison of \modelname and baseline ensemble methods.}
  \label{fig:system_efficiency_figure}
    \vspace{-2em}
\end{figure*}

% \textbf{Time Efficiency Analysis.}
% We evaluate the inference latency of sequential and near-parallel execution strategies under different ensemble sizes, as shown in Figure~\ref{fig:inference_efficiency}. The results show that sequential inference latency increases almost linearly with the number of models, making scaling expensive in a real-world setting. In contrast, our near-parallel execution reduces the total inference cost to 12.1 seconds, achieving a 47\% reduction in latency compared to sequential execution. This demonstrates that near-parallelization enables \modelname to scale multiple models efficiently while maintaining practical inference speed.

\textbf{Efficiency and Cost Analysis.}
We evaluate end-to-end wall-clock time, per-token latency, and peak GPU memory during both training and inference, with results shown in Figure~\ref{fig:system_efficiency_figure}. \modelname (7B+7B, near-parallel, 2 GPUs) achieves an end-to-end latency of 13.75 seconds, significantly outperforming ensemble baselines such as UNITE (22.66 s), VOTE (22.61 s), and T-Copilot (17.21 s), and it remains competitive even when restricted to a single GPU. The per-token latency of \modelname (0.075 s/token) is also lower than that of VOTE (0.127 s/token), UNITE (0.128 s/token), and T-Copilot (0.097 s/token), reflecting improved throughput. In terms of memory usage, \modelname reaches 33.18\% peak training memory and 20.72\% peak inference memory per GPU in the 2-GPU configuration, which is comparable to or below the levels of other ensemble systems. The 1-GPU results follow a similar trend. The cost introduced by cross-model state passing is minimal 0.0009 s on 2 GPUs and 0.0002 s on 1 GPU and has negligible impact on the overall latency. Taken together, these findings show that \modelname achieves a more favorable combination of speed, memory efficiency, and coordination cost than existing ensemble approaches, while still providing strong accuracy gains.
All efficiency measurements were conducted on an NVIDIA H800 GPU cluster using the AQuA dataset, with identical decoding settings across all methods. 
A theoretical analysis of inference-time behavior under different execution strategies is provided in Appendix~\ref{app:inferenceTimeEfficiencyAnalysis}.

\textbf{Comparison under equal parameter budgets.}
We compare \modelname with single-model baselines of equivalent parameter counts under the supervised fine-tuning (SFT) setting. As shown in Table~\ref{tab:main_result}, \modelname built from two Qwen2.5-3B models (6B total) reaches the same level of commonsense performance as the 7B model and achieves a higher arithmetic score, improving the average from 77.5 to 78.0, while using 1B fewer parameters.
Under a 14B budget, \modelname constructed from two Qwen2.5-7B models also shows consistent advantages. On HellaSwag, performance increases from 92.9 to 95.1; on SVAMP, from 87.4 to 88.8; and the overall commonsense average rises from 85.1 to 85.6. Arithmetic reasoning similarly improves from 80.1 to 80.2.
In terms of efficiency, per-token latency decreases from 98 ms to 75 ms, corresponding to a 23\% speedup under the same parameter budget. Overall, \modelname provides both higher accuracy and faster inference than single LLMs with equivalent total size.

\textbf{Weakly Heterogeneous Model Ensemble.} We further evaluate \modelname under a weakly heterogeneous configuration by integrating two models from the same family but belonging to different series. In practice, integrating models from different families is difficult because their tokenizers are incompatible; even within a single family, different parameter sizes may adopt different vocabularies (e.g., Qwen2.5-7B and Qwen2.5-3B). Such tokenizer mismatches lead to inconsistent token boundaries during decoding, preventing \modelname from performing accurate layer-wise residual correction. Based on this, we focus on the compatible model pair of Llama3.1-8B and Llama3.2-3B, which maintain consistency in both tokenizer and architectural specifications.
As shown in Table~\ref{tab:main_result}, the weakly heterogeneous ensemble (8B + 3B) improves the average commonsense score from 80.0 to 83.7 and raises the arithmetic average from 68.2 to 69.1 compared with the stronger 8B model alone. This accuracy gain comes with only a minimal increase in computational cost: the per-token latency rises slightly from 0.038 s to 0.049 s, indicating that the ensemble adds only marginal overhead. These results show that even when combining models of different series and scales, \modelname continues to provide consistent benefits in both accuracy and efficiency. This demonstrates that our boosting framework extends beyond strictly homogeneous ensembles and offers a viable approach for integrating diverse model variants within the same family.

\begin{wraptable}{r}{0.4\textwidth} % 稍微调窄一点宽度
    \vspace{-20pt}
    \centering
    \caption{Performance on CCAD.}
    \label{exp:dianxindataset}
    % 使用 \small 或 \footnotesize 调整基准字体大小
    % 注意：如果用了 resizebox，这里的字体大小影响不大，
    % 建议先去掉 resizebox 或者把表格内容做得更紧凑
    \resizebox{\linewidth}{!}{ 
        % \begin{table}[H] % H = 强制当前位置 (需要 \usepackage{float})
% \caption{Accuracy (\%) on \dianxindataset.}
% \label{exp:dianxindataset}
% \resizebox{\linewidth}{!}{
% \begin{tabular}{l|l|l}
%     \toprule
%     Model & Params & \dianxindatasetshort   \\ 
%     \midrule
%     \multicolumn{3}{c}{\textbf{Llama-3.1-8B}} \\
%     \midrule
%     Llama & $1\times8B$ & 52.0\\
%     VOTE & $2\times8B$ & 53.0\\ 
%     UNITE & $2\times8B$ & 52.5\\ 
%     \modelname & $2\times8B$ & \textbf{55.0}\\
%     \midrule
%     \multicolumn{3}{c}{\textbf{Qwen-2.5-7B}} \\
%     \midrule
%     Qwen & $1\times7B$ & 41.0 \\
%     VOTE & $2\times7B$ &  43.0 \\ 
%     UNITE & $2\times7B$ & 41.5 \\ 
%     \modelname & $2\times7B$ & \textbf{45.5}\\
%     \bottomrule
% \end{tabular}
% }

% \end{table}

\begin{tabular}{@{}l c c@{}} % @{} 去除左右多余留白，lcc 表示左对齐、居中、居中
    \toprule
    Model & Params & Acc. (\%) \\ 
    \midrule
    \multicolumn{3}{c}{\textit{Llama-3.1-8B Base}} \\
    \midrule
    Single & 8B & 52.0 \\
    VOTE & $2\times8B$ & 53.0 \\ 
    UNITE & $2\times8B$ & 52.5 \\ 
    \rowcolor{gray!14} % 添加浅灰色背景突出你的模型
    \modelname & $2\times8B$ & \textbf{55.0} \\
    \midrule
    \midrule
    \multicolumn{3}{c}{\textit{Qwen-2.5-7B Base}} \\
    \midrule
    Single & 7B & 41.0 \\
    VOTE & $2\times7B$ & 43.0 \\ 
    UNITE & $2\times7B$ & 41.5 \\ 
    \rowcolor{gray!14}
    \modelname & $2\times7B$ & \textbf{45.5} \\
    \bottomrule
\end{tabular}
    }
    \vspace{-10pt}
\end{wraptable}
\textbf{Performance on \dianxindatasetshort.} 
As detailed in Table~\ref{exp:dianxindataset}, \textsc{\modelname} establishes a new state-of-the-art on \dianxindataset, exhibiting remarkable robustness across the Llama and Qwen model families. 
Compared to traditional ensemble strategies like VOTE and UNITE, our approach yields consistent improvements of 3.0\% (Llama-3.1-8B) and 2.5\% (Qwen-2.5-7B), highlighting its ability to effectively aggregate weak learners. 
Crucially, the advantages of \textsc{\modelname} are even more pronounced in deployment settings. 
In large-scale end-to-end toolchain scheduling evaluations conducted at \dianxinlab, our method achieves a significant 5\% gain over existing solutions. 
This result serves as strong evidence of its scalability and practical utility in optimizing complex, industrial-grade production pipelines.
\section{Conclusion}
% We propose \modelname, a novel ensemble learning framework for LLMs with boosting style. It leverages a cross-model attention to enable subsequent models to use predecessor models’ intermediate hidden states. We further introduce a near-parallel inference paradigm that significantly improves efficiency. Extensive experiments show that \modelname outperforms baselines like VOTE, effectively enhancing the model’s reasoning ability, and significantly accelerating inference efficiency compared to traditional ensembles.More discussions are provided in Appendix \ref{appendix:limitation}
We propose \modelname, a novel boosting-style ensemble learning framework for LLMs. It leverages cross-model attention, enabling subsequent models to utilize the intermediate hidden states of their predecessors. We further introduce a near-parallel inference paradigm that substantially improves efficiency. Extensive experiments demonstrate that \modelname outperforms baselines such as VOTE, effectively enhancing reasoning ability and significantly accelerating inference compared to traditional ensembles. Additional discussions are provided in Appendix \ref{appendix:limitation}.

\textbf{Future work.} In future work, we plan to extend our framework to multi-agent settings, where different agents can specialize in reasoning, planning, or tool usage. We will further explore adaptive routing mechanisms that dynamically select agents or reasoning paths based on task context and system states, enabling faster inference and more efficient resource utilization. Combined with speculative reasoning, such routing is expected to reduce latency and improve scalability in large-scale deployments.

\clearpage
\bibliographystyle{iclr2026_conference}
\bibliography{iclr2026_conference}

\clearpage

\clearpage

\appendix
\textbf{\Large Appendix}

\addtocontents{toc}{\protect\setcounter{tocdepth}{2}}

\tableofcontents

\newpage
\section{Symbols used in the text}

\begin{table}[htbp]
\centering
\caption{LLMBoost Symbols}
\label{tab:llmboost_symbols}
\renewcommand{\arraystretch}{1.2}
\begin{tabular}{@{}p{0.9in}p{3.5in}@{}}
\toprule
$\displaystyle v$ & Vocabulary dimension index \\
$\displaystyle A_i$ & The sampling space of the model  The sampling space of the model $m_i$ \\
$\displaystyle \mathcal{A}_i$ & The sampling space of the ensemble model $\mathcal{M}_i$ \\
$\displaystyle \mathcal{D}$ & Training dataset \\
$\displaystyle d_k$ & Dimension of key matrix \\
$\displaystyle e^{(i)}_{t, v}$ & Residual of $\mathcal{M}_i$ at vocab dimension $v$, t-th token \\
$\displaystyle \mathbb{E}$ & Expectation operator \\
$\displaystyle \mathcal{M}_i$ & Ensemble model at stage $i$ \\
$\displaystyle g$ & Global shared pool \\
$\displaystyle h^{i}_{l,t}$ & Hidden state of $i$-th model,layer $l$, step $t$ \\
$\displaystyle g^{(i)}_{t,v}$ & Linear term of the model $m_i$ in Taylor expansion of softmax \\
$\displaystyle K^{i}_{l,t}$ & Key matrix of $i$-th model,layer $l$, step $t$ \\
$\displaystyle \mathcal{L}_i$ & Loss of $i$-th model \\
$\displaystyle \lambda_i$ & Scaling coefficient for logits fusion \\
$\displaystyle M_i$ & Initial untuned base model \\
$\displaystyle \tilde{M}_i$ & $i$-th model after LoRA fine-tuning \\
$\displaystyle \mathcal{M}$ & Set of base models \\
$\displaystyle \text{MSE}$ & Total Mean squared error\\
$\displaystyle p^{(i)}_t$ & Probability distribution of $i$-th model, t-th token \\
$\displaystyle P^{(i)}_t$ & Probability distribution of $\mathcal{M}_i$, t-th token \\
$\displaystyle Q^{i}$ & Query matrix of $i$-th model \\
$\displaystyle \mathcal{V}$ & Vocabulary set \\
$\displaystyle W_Q, W_K, W_V$ & Projection weight matrices \\
$\displaystyle x$ & Input sequence \\
$\displaystyle y^{*}$ & Ground-truth sequence \\
$\displaystyle \hat{y}_t$ & Token generated at step $t$ \\
$\displaystyle z^{(i)}_t$ & Raw logits of $i$-th model, t-th token\\
$\hat{z}^{(i)}_t$ & Accumulated logits for ensemble $\mathcal{M}_{i}$ at t-th token \\
$\displaystyle \texttt{EOS}$ & End-of-sentence token \\
$\displaystyle \phi$ & LoRA configuration\\
$\displaystyle \theta_i$ & Trainable parameters of $i$-th model \\
$\Theta_{i-1}$  & Parameters of the ensemble $\mathcal{M}_{i-1}$ \\
$X_\tau$ & the input sequence\\
$Y_\tau$ &the corresponding target sequence\\
\bottomrule
\end{tabular}
\end{table}

\section{Broader Impact and Future Applications \& Limitations}
%\section{Features and Future Applications \& Limitation}
\label{appendix:limitation}
\textbf{Broader Impact and Future Applications} \modelname has practical value for real-world LLM deployment. For multi-GPU tasks, it enables distributed ensemble training/inference—using layer-wise parallelism and lightweight LoRA to boost model performance while reducing the computational overhead of full-model replication. For AI agents, it integrates directly into existing systems without modifying the agents themselves, helping them improve the reliability of cloud tasks such as tool scheduling and batch data verification via its error-correction ability, without disrupting workflows. This balances performance and deployment simplicity, fitting scenarios where both matter.

\textbf{Limitations} Despite its effectiveness, \modelname currently has a key cross-model compatibility limitation: it relies on consistent structural design and hidden state formats of ensemble sub-models, restricting it to homogeneous families like LLaMA-3 or Qwen-2.5 series. For heterogeneous ensembles combining models such as LLaMA-3 and Qwen-2.5, discrepancies in hidden state dimensions, feature encoding logics, and layer-wise designs disable the residual cross-attention mechanism. This disruption impairs both the chained training paradigm—where subsequent models lack reliable error-correction signals—and the near-parallel inference paradigm—where cross-model hidden state transmission fails due to structural mismatches—ultimately making \modelname inapplicable to such setups. This limitation points to future work: developing a cross-model adaptation module to bridge structural and representational gaps via targeted design strategies. Such a module would enable \modelname to leverage the complementary strengths of diverse model families and expand its applicability.
﻿

\section{Datasets \& Baselines}
\subsection{DATASETS INTRODUCTION}\label{appendix:arithme}
% We evaluate our method on two main task types—commonsense reasoning and arithmetic reasoning—along with a domain-specific cloud service dataset (CCAT). 
Our evaluation covers two primary task types—commonsense reasoning and arithmetic reasoning—and further extends to a domain-specific cloud service dataset (\dianxindatasetshort).

\textbf{Commonsense Reasoning Datasets} We select six open-ended multiple-choice QA tasks to assess commonsense reasoning ability:
\label{appendix:datasets}
\begin{itemize}[leftmargin=*]
    \item \textbf{PIQA} \citep{bisk2020piqa}: A dataset for physical interaction commonsense reasoning, where models select the more reasonable solution from 2 options for daily physical tasks.

    \item \textbf{WinoGrande (WinoG.)} \citep{sakaguchi2021winogrande}: A coreference resolution dataset with sentences containing ambiguous pronouns, requiring models to identify the correct referent via commonsense.

    \item \textbf{HellaSwag (HellaS.)} \citep{zellers2019hellaswag}: A scenario continuation dataset, where models select the most logically consistent ending from 4 options for a given partial daily scenario.

    \item \textbf{BoolQ} \citep{clark2019boolq}: A fact-based commonsense dataset with yes/no questions derived from web texts, requiring models to output "True" or "False" based on commonsense and text information.

    \item \textbf{SIQA} \citep{sap2019socialiqa}: A social commonsense dataset presenting daily social scenarios, where models choose the appropriate response from 2 options to assess social etiquette understanding.

    \item \textbf{Openbook QA (OBQA)} \citep{mihaylov2018can}: A structured commonsense dataset linked to a "open book" of basic scientific facts, with 4-option multiple-choice questions requiring models to apply the book’s facts to answer.
\end{itemize}

\textbf{Arithmetic Reasoning Datasets}We use four open-ended math problem-solving datasets spanning multiple mathematical domains to evaluate arithmetic reasoning ability:

\begin{itemize}[leftmargin=*]
    \item \textbf{AQuA} \citep{ling2017program}: An algebraic reasoning dataset with middle-school algebra problems, requiring models to construct equations and output numerical answers.

    \item \textbf{GSM8K} \citep{cobbe2021training}: A multi-step arithmetic dataset with elementary math word problems, often requiring step-by-step reasoning and numerical answers.

    \item \textbf{MAWPS} \citep{koncel2016mawps}: A diverse arithmetic dataset with integer, decimal, and fraction word problems, where models output numerical answers.

    \item \textbf{SVAMP} \citep{patel2021nlp}: A single-variable arithmetic dataset with word problems convertible to single-variable equations, requiring models to output numerical answers.
\end{itemize}

In both commonsense and arithmetic reasoning experiments, we follow the setup of \cite{hu2023llm}. For commonsense reasoning, models are fine-tuned on \textit{Commonsense170K}, built from six benchmark training sets, and evaluated on their official test sets. For arithmetic reasoning, models are fine-tuned on \textit{Math10K}, derived from four benchmarks, and evaluated likewise. All commonsense instances use zero-shot prompts in both fine-tuning and evaluation, while arithmetic data follow the preprocessing protocol of \cite{wu2024reft} to prevent data leakage.

\textbf{\dianxindataset(\dianxindatasetshort)} A domain-specific dataset for cloud service tool scheduling, derived from \dianxinlab’s real user interaction logs. Each sample includes a predefined tool library, a user inquiry, and requires models to select the most appropriate tools from the library to address the inquiry.

% For both commonsense reasoning and arithmetic reasoning experiments, we follow the experimental setup from \cite{hu2023llm}: we fine-tune our models and baseline models on combined training datasets—Commonsense170K, constructed by sampling and integrating the training sets of the six commonsense reasoning datasets, for commonsense reasoning, and Math10K, a combination of the four arithmetic reasoning datasets, for arithmetic reasoning—with each task’s individual dataset test sets used for evaluation; zero-shot input prompts are applied to all fine-tuning and testing data instances for commonsense reasoning, while we adopt the data preprocessing setup from \cite{wu2024reft} for arithmetic reasoning to avoid potential training data leakage.
\label{appendix:baselines}
\subsection{Baselines}
\label{app:baselines}
We further analyze the improvements of \textsc{\modelname} over representative ensemble methods:
\begin{itemize}[leftmargin=*]
    \item \textbf{Vote} \citep{li2024more}: The vote-based method for large model ensembling aggregates predictions from multiple individual models by having them vote on the final output.It leverages the complementary strengths of different models to reduce individual model biases and improve overall prediction robustness.
    \item \textbf{UNITE}\citep{yao2024determine}: UNITE is a novel LLM ensembling method that efficiently combines models by focusing on the union of top-k tokens from each model, avoiding full vocabulary alignment and reducing computational overhead.It incorporates a model selection strategy identifying compatible models, leveraging their complementary advantages to enhance performance across multiple benchmarks.
    \item \textbf{T-copilot}\citep{zou2025transformer}: T-copilot is a method that adapts large language models to downstream tasks by retaining the model’s own learning signals, using a Pilot-Copilot framework with joint training and fused inference.It tracks the Pilot model’s errors via a Mistake Log, which the Copilot uses to rectify the Pilot’s logits for better generation.
    % \item \textbf{EnsemW2S}\cite{agrawal2025ensemw2s}:
\end{itemize}

\section{Near-Parallel Decoding Psudo Algorithm}
\label{app:Near_Parallel_Algorithm}
\begin{algorithm}[!htbp]
\caption{Near-Parallel Decoding Psudo Algorithm} 
\label{alg:parallel_inference_simple} 
\SetAlgoLined 
\LinesNumbered 
\KwIn{Model list $\mathcal{\tilde{M}}= \{\tilde{m}_0(\cdot; \theta_0), \tilde{m}_1(\cdot; \theta_1), \dots, \tilde{m}_n(\cdot; \theta_n)\}$; input $x$; max Tokens $T$; global shared pool $g$;hidden states $h^{(i)}_{l,t}$ of the $i$-th model at layer $l$ for token $t$.} 
\KwOut{Final output $\{\hat{y}_{0} \dots \hat{y}_{t}\}$}  
Initialize shared pool $g \gets \emptyset$\; 
Run $\tilde{m}_0$ in main process, others $\tilde{m}_i\ (i\!\geq\!1)$ in parallel threads\; 
\For{$t=1$ \KwTo $T$}{
    % \colorbox{gray!20}{$\triangledown$ lines 5-9 Models $0 \!-\! k$ execute simultaneously} \\
    \ForPar{$i=1$ \KwTo $n$}{
        \For{$l=1$ \KwTo $L$}{         
            Each model $\tilde{m}_i$ waits for $h^{(i-1)}_{l,t}$ in $g$\; 
            Compute hidden state $h^{(i)}_{l,t}$ at layer $l$ with available inputs\;         
            Immediately push $h^{(i)}_{l,t}$ into $g$ for downstream use\;  
        } 
    }
    Each model $\tilde{m}_i$ produces t-th logits $z^{(i)}_t$\;  
    Fuse logits $z^0_t$ according to Eq.~(\ref{equ:logitsFusion})\;  
    Decode token $\hat{y}_{t} \gets \mathrm{Argmax}(\tilde{z}^0_t)$ \;  
    \If{$\hat{y}_{t}=\texttt{EOS}$}{break} 
} 
\KwRet{$\{\hat{y}_{0} \dots \hat{y}_{t}\}$} 
\end{algorithm} 
The pseudo code in Algorithm~\ref{alg:parallel_inference_simple} illustrates the overall workflow of our near-parallel decoding strategy. 
The key idea is to allow multiple models to decode tokens in parallel while maintaining synchronization through a lightweight communication mechanism. 
By progressively sharing intermediate hidden states, the framework achieves both efficiency and consistency in multi-model inference.

Specifically, the global shared pool $g$ is implemented as a thread-safe queue that stores hidden states produced by each model at every layer. 
Once a model finishes computing a hidden state $h^{(i)}_{l,t}$, it immediately pushes the result into $g$, where it can be consumed by downstream models. 
This ensures that information flows smoothly across different models without requiring full-sequence synchronization.

In practice, we employ multi-threading for parallel execution, where each auxiliary model $\tilde{m}_i$ runs in a separate thread. 
To guarantee correctness, a lightweight lock mechanism is introduced so that models can safely wait for the availability of required hidden states in $g$ before continuing computation. 
This design provides fine-grained synchronization with minimal overhead, enabling near-parallel decoding while avoiding race conditions.

\newpage
\section{Fine-tuning Dataset Template}

\begin{table}[H]
    \centering
    \small
    \caption{Examples of dataset templates used in \modelname.}
    \label{tab:dataset_template}
    \resizebox{\linewidth}{!}{%
    \begin{tabular}{l|p{0.9\linewidth}}
    \toprule
    \textbf{Task Type} & \textbf{Template} \\
    \midrule

    \multirow{15}{*}{\textbf{Commonsense Reasoning}} &
    \textbf{Input:} Please choose the correct solution to the question: To fight Ivan Drago in Rocky for sega master system.\par
    Solution1: Drago isn't in this game because it was released before Rocky IV.\par
    Solution2: You have to defeat Apollo Creed and Clubber Lang first.\par
    Answer format: solution1/solution2
    \par\vspace{1ex}
    \textbf{Answer:} the correct answer is solution2. \\
    \cmidrule{2-2}

    &
    \textbf{Input:} Please choose the correct answer to fill in the blank to complete the given sentence: Sarah was a much better surgeon than Maria so () always got the harder cases.\par
    Option1: Sarah\par
    Option2: Maria\par
    Answer format: option1/option2
    \par\vspace{1ex}
    \textbf{Answer:} the correct answer is option1. \\
    \cmidrule{2-2}

    &
    \textbf{Input:} Please answer the following question with true or false: is there a difference between hydroxyzine hcl and hydroxyzine pam?\par
    Answer format: true/false
    \par\vspace{1ex}
    \textbf{Answer:} the correct answer is true. \\
    \midrule

    \multirow{19}{*}{\textbf{Arithmetic Reasoning}} &
    \textbf{Input:} In a school there are 569 girls and 236 boys. How many more girls than boys does the school have?
    \par\vspace{1ex}
    \textbf{Answer:}
    There are 569 girls and 236 boys. The difference between them is $569 - 236 = 333$. The school has 333 more girls than boys. \\
    \cmidrule{2-2}

    &
    \textbf{Input:} A study reported that in a random sampling of 100 women over the age of 35, 8 of the women were married two or more times. Based on the study results, how many women in a group of 5000 women over the age of 35 would likely be married two or more times?
    \par\vspace{1ex}
    \textbf{Answer:} $x = (8/100) \times 5000 = 400$. \\
    \cmidrule{2-2}

    &
    \textbf{Input:} Twenty kids went out on a school trip. They were divided into two equal groups of girls and boys. The boys brought back 60 shells each. If the girls brought the number of shells brought by the boys plus four times as many, how many shells were brought by each girl?
    \par\vspace{1ex}
    \textbf{Answer:} $60 + 4 \times 60 = 300$. The answer is 300. \\
    \midrule

    \multirow{18}{*}{\textbf{\dianxindatasetshort}} &
    \textbf{Input:} Cloud agent can use the following tools:\par
    (1) \textless name\textgreater QueryCloudServerCount\textless/name\textgreater\par
    \textless describe\textgreater query the number of cloud servers under the user's account\textless/describe\textgreater\par
    (2) \textless name\textgreater QueryUserQuota\textless/name\textgreater\par
    \textless describe\textgreater query global and resource pool quotas\textless/describe\textgreater\par
    (3) \textless name\textgreater QueryCloudServerInformation\textless/name\textgreater\par
    \textless describe\textgreater query cloud server specifications\textless/describe\textgreater\par
    (4) \textless name\textgreater QueryCloudResourceList\textless/name\textgreater\par
    \textless describe\textgreater query all cloud resource names\textless/describe\textgreater\par

    When the cloud agent receives an inquiry: \emph{Can I create 2 more cloud hosts?} Which tool is most likely to be called?\par
    (A) QueryCloudServerCount and QueryUserQuota\par
    (B) QueryUserQuota\par
    (C) QueryCloudServerCount and QueryCloudServerInformation\par
    (D) QueryCloudServerInformation and QueryCloudResourceList
    \par\vspace{1ex}
    \textbf{Answer:} A. \\
    \bottomrule

    \end{tabular}}
\end{table}

In Table \ref{tab:dataset_template}, we provide examples of data instances for each task mentioned above during model fine-tuning. All experiments are conducted in the zero-shot setting to better facilitate model-wise evaluation using  accuracy.

\section{Inference Time Efficiency Analysis}
\label{app:inferenceTimeEfficiencyAnalysis}

We analyze the inference latency under different execution strategies.

\paragraph{Sequential execution.}
In the naive setting, each model $\tilde{M}_i$ must finish all $L$ layers
before the next model starts, with per-layer compute cost $c$ . 
% With per-layer compute cost $c$ and output length $T$, 
The total latency is
\begin{equation}
  T_{\text{sequential}} \;\approx\; (n+1) \cdot L \cdot c.
\end{equation}

\paragraph{Parallel execution with inter-model dependency.}
We consider $k$ models, each with $l$ layers, executed on $g$ GPUs in a round-robin assignment. 
Let the execution of one layer on a dedicated GPU take time $c$. Due to inter-model dependencies, the $i$-th layer of model $m$ can only start after the $i$-th layer of model $m-1$ and the $(i-1)$-th layer of model $m$ have completed. 
Hence, the computation graph forms a $k\times l$ grid with precedence constraints along both the horizontal and vertical directions. 

Define an anti-diagonal index $s = m+i$, so that all tasks $\{(m,i): m+i=s\}$ can be executed in parallel once their predecessors complete. 
The number of tasks on the $s$-th anti-diagonal is
\begin{equation}
d_s \;=\; \min(k,\,s-1) - \max(1,\,s-l) + 1.
\end{equation}
These values increase from $1$ to $w=\min(k,l)$, stay constant at $w$ for $u-w+1$ steps where $u=\max(k,l)$, and then decrease symmetrically back to $1$. 
Geometrically, the execution order thus corresponds to a \emph{parallelogram}, which can be decomposed into two symmetric triangular parts and a central rectangular part.

On the $s$-th anti-diagonal, round-robin assignment ensures that tasks are distributed across GPUs as evenly as possible. 
Therefore, the maximum number of tasks assigned to a single GPU is exactly $\lceil d_s/g \rceil$, and the completion time of this anti-diagonal is 
\begin{equation}
T_s \;=\; c \cdot \Big\lceil \tfrac{d_s}{g}\Big\rceil.
\end{equation}
Summing over all anti-diagonals, the total execution time is
\begin{equation}
T_{\text{parallel}} \;=\; c\sum_{s=2}^{k+l}\Big\lceil \tfrac{d_s}{g}\Big\rceil.
\end{equation}
By decomposing the parallelogram into two triangles and one rectangle, we obtain the closed-form expression
\begin{equation}
T_{\text{parallel}} \;=\; 2c \sum_{t=1}^{w-1} \Big\lceil \tfrac{t}{g}\Big\rceil \;+\; (u-w+1)c \Big\lceil \tfrac{w}{g}\Big\rceil + \delta,
\end{equation}
where $w=\min(k,l)$, $u=\max(k,l)$ and $\delta$ is communication overhead.

\paragraph{Speedup over sequential execution.}
The theoretical speedup compared to the naive sequential strategy is
\begin{equation}
  \mathrm{Speedup}
  \;=\;
  \frac{T_{\text{sequential}}}{T_{\text{parallel}}}
  \;=\;
  \frac{(n + 1) \cdot L}{\displaystyle
  2 \sum_{t=1}^{w-1}\Big\lceil \tfrac{t}{g}\Big\rceil
  + (u-w+1)\Big\lceil \tfrac{w}{g}\Big\rceil + \delta}.
\end{equation}
This expression clearly shows the achievable acceleration:
(i) if $g=1$, the speedup is $1$, as expected;
(ii) if $g \ge w$, the parallel latency reduces to $c(k+l-1)$, and the speedup approaches $\tfrac{kl}{k+l-1}$, which scales almost linearly in $\min(k,l)$;
(iii) for intermediate $g$, the speedup smoothly interpolates between these two extremes, depending on how many anti-diagonal tasks can be executed simultaneously.

\section{Additional Experiments}
\label{appendix: additionalexperiments}

\subsection{Implementation Details.}
\label{app: implementationDetails}
Tables \ref{tab:hyper_commonsense} and \ref{tab:hyper_arithmetic} summarize the main training details for reproducibility. In our model, the fusion coefficient of cross-model attention and the coefficient for error suppression are both set to 0.1. All reported results can be reproduced with random seeds 1, 2, and 3.
\begin{table}[!ht]
    \centering
    \caption{Unified hyperparameter configuration of \modelname for LLaMA-3 and Qwen-2.5 series models on \textbf{Commonsense Reasoning} tasks.}
    \label{tab:hyper_commonsense}
    \resizebox{0.8\linewidth}{!}{%
    \begin{tabular}{l|cccc}
    \toprule
    \multirow{2}{*}{\textbf{Hyperparameters}} & \multicolumn{2}{c}{\textbf{Llama Models}} & \multicolumn{2}{c}{\textbf{Qwen Models}} \\
    \cmidrule(lr){2-3} \cmidrule(lr){4-5}
     & LLaMA-3.2-3B & LLaMA-3.1-8B & Qwen-2.5-3B & Qwen-2.5-7B \\

    \midrule
    \rowcolor{gray!20} \multicolumn{5}{l}{\textit{\textbf{Fine-tuning Configurations}}} \\
    Epochs & 2 & 2 & 2 & 2 \\
    Batch Size & 16 & 16 & 16 & 16 \\
    Micro Batch Size & 4 & 4 & 4 & 4 \\
    Cut Off Length & 256 & 256 & 256 & 256 \\
    Maximum Learning Rate & $2e^{-4}$ & $2e^{-4}$ & $2e^{-4}$ & $2e^{-4}$ \\
    Learning Rate Scheduler & Cosine & Cosine & Cosine & Cosine \\
    Optimizer & AdamW & AdamW & AdamW & AdamW \\
    Warmup Steps & 200 & 200 & 200 & 200 \\
    Weight Decay & 0.00 & 0.00 & 0.00 & 0.00 \\
    \midrule
    \rowcolor{gray!20} \multicolumn{5}{l}{\textit{\textbf{LoRA Configurations}}} \\
    Rank $r$ & 32 & 32 & 32 & 32 \\
    LoRA Alpha & 64 & 64 & 64 & 64 \\
    LoRA Dropout & 0.05 & 0.05 & 0.05 & 0.05 \\
    \midrule
    \rowcolor{gray!20} \multicolumn{5}{l}{\textit{\textbf{Inference Configurations}}} \\
    Temperature & \multicolumn{4}{c}{0.1} \\
    Top p & \multicolumn{4}{c}{0.95} \\
    Top k & \multicolumn{4}{c}{40} \\
    Maximum New Tokens & \multicolumn{4}{c}{64} \\
    \bottomrule
    \end{tabular}%
    }
\end{table}

\begin{table}[!ht]
    \centering
    \caption{Unified hyperparameter configuration of \modelname for LLaMA-3 and Qwen-2.5 series models on \textbf{Arithmetic Reasoning} tasks.}
    \label{tab:hyper_arithmetic}
    \resizebox{0.8\linewidth}{!}{%
    \begin{tabular}{l|cccc}
    \toprule
    \multirow{2}{*}{\textbf{Hyperparameters}} & \multicolumn{2}{c}{\textbf{Llama Models}} & \multicolumn{2}{c}{\textbf{Qwen Models}} \\
    \cmidrule(lr){2-3} \cmidrule(lr){4-5}
     & LLaMA-3.2-3B & LLaMA-3.1-8B & Qwen-2.5-3B & Qwen-2.5-7B \\
    \midrule
    \rowcolor{gray!20} \multicolumn{5}{l}{\textit{\textbf{Fine-tuning Configurations}}} \\
    Epochs & 2 & 2 & 2 & 2 \\
    Batch Size & 16 & 16 & 16 & 16 \\
    Micro Batch Size & 4 & 4 & 4 & 4 \\
    Cut Off Length & 256 & 256 & 256 & 256 \\
    Maximum Learning Rate & $2e^{-4}$ & $2e^{-4}$ & $2e^{-4}$ & $2e^{-4}$ \\
    Learning Rate Scheduler & Cosine & Cosine & Cosine & Cosine \\
    Optimizer & AdamW & AdamW & AdamW & AdamW \\
    Warmup Steps & 100 & 100 & 100 & 100 \\
    Weight Decay & 0.00 & 0.00 & 0.00 & 0.00 \\
    \midrule
    \rowcolor{gray!20} \multicolumn{5}{l}{\textit{\textbf{LoRA Configurations}}} \\
    Rank $r$ & 32 & 32 & 32 & 32 \\
    LoRA Alpha & 64 & 64 & 64 & 64 \\
    LoRA Dropout & 0.05 & 0.05 & 0.05 & 0.05 \\
    \midrule
    \rowcolor{gray!20} \multicolumn{5}{l}{\textit{\textbf{Inference Configurations}}} \\
    Temperature & \multicolumn{4}{c}{0.1} \\
    Top p & \multicolumn{4}{c}{0.95} \\
    Top k & \multicolumn{4}{c}{40} \\
    Maximum New Tokens & \multicolumn{4}{c}{256} \\
    \bottomrule
    \end{tabular}%
    }
\end{table}

\subsection{Ablation Study}
\begin{table}[!ht]
\centering
\caption{Ablation study of key components in Ablation study of key components in \textsc{\modelname}, evaluated on mathematical reasoning benchmarks (, evaluated on mathematical reasoning benchmarks (SVAMP, , MAWPS). ``wo E.S.O.'' removes the Error-Suppression Objective; ``wo C.A.'' disables Cross-Model Attention; ``wo ). ``wo E.S.O.'' removes the Error-Suppression Objective; ``wo C.A.'' disables Cross-Model Attention; ``wo $top$--$k$'' ablates the Key-Token Backward Pass (with '' ablates the Key-Token Backward Pass (with $k=2$). ). \modelname consistently outperforms all variants, validating the contribution of each module. consistently outperforms all variants, validating the contribution of each module.}
\label{exp:ablationStudy}
\begin{tabular}{l|lll}
    \toprule
    MODEL TYPE & MAWPS & SVAMP & Avg.  \\ 
    \midrule
    \multicolumn{4}{c}{\textbf{Qwen-2.5-7B as base model}} \\
    \midrule
    \modelname ($2\times7B$) & 88.8 & 94.1 & 91.5\\ 
    wo E.S.O. & 87.3 & 93.3 & 90.3 \\
    wo C.A. & 87.4 & 92.4 & 89.9 \\
    wo $top$-$k$ & 87.2 & 93.3 & 90.3\\
    \bottomrule
\end{tabular}
\end{table}
To better understand the contribution of each component in our method, we conduct ablation experiments to verify the effectiveness of different modules. Unless otherwise stated, all experiments are performed under the same settings as in the main results.

We first investigate the effectiveness of the key components in our method. Specifically, we analyze the impact of the Error-Suppression Objective(E.S.O.) and cross-model attention(C.A.). We compare \modelname with its ablated variants to understand the contribution of each component.

\subsection{Effect of $\lambda$ Choice}
\begin{table}
\centering
\caption{Sensitivity analysis of the weighting coefficient(s) $\lambda$ in \modelname. Performance (\%) is evaluated across arithmetic reasoning benchmarks (AQuA, MAWPS, SVAMP) under matched parameter scales. The upper block explores $\lambda_1 \in [0.1, 1.0]$ for the \modelname ($2 \times 8B$), while the lower block studies combinations of $\lambda_1, \lambda_2$ for the \modelname ($3 \times 8B$).}
\label{exp:lambdaChoice}
\begin{tabular}{l|c|c c c c}
    \toprule
    Params & $\lambda$ & AQuA & MAWPS & SVAMP & Avg. \\ 
    \midrule
    \multicolumn{6}{c}{\textbf{Llama-3.1-8B as base model}}\\
    \midrule
    \multirow{10}{*}{$2 \times 8B$}
    & 0.1 & 45.7 & 89.4 & 78.9 & 71.3 \\ 
    & 0.2 & 46.4 & 90.1 & 80.0 & 72.2 \\ 
    & 0.3 & 46.8 & 90.1 & 80.1 & \textbf{72.3} \\
    & 0.4 & 43.9 & 90.1 & 80.1 & 71.4 \\ 
    & 0.5 & 46.3 & 89.3 & 79.8 & 71.8 \\ 
    & 0.6 & 46.1 & 90.3 & 79.6 & 72.0 \\
    & 0.7 & 43.9 & 89.7 & 80.0 & 71.2 \\ 
    & 0.8 & 45.3 & 90.7 & 79.9 & 72.0 \\ 
    & 0.9 & 45.7 & 89.9 & 79.5 & 71.7 \\
    & 1.0 & 45.6 & 89.8 & 79.2 & 71.5 \\
    \midrule
    \multirow{3}{*}{$3 \times 8B$}
    & 0.3, 0.1 & 46.9 & 90.7 & 80.4 & 72.7 \\ 
    & 0.3, 0.2 & 46.6 & 92.1 & 80.8 & 73.2 \\ 
    & 0.3, 0.3 & 47.6 & 92.0 & 81.3 & \textbf{73.6} \\
    \bottomrule
\end{tabular}
\end{table}

In addition to ablation studies, we conduct a detailed sensitivity analysis of the weighting coefficient(s) $\lambda$, which control the balance between different loss components in \modelname. As shown in Table~\ref{exp:lambdaChoice}, for the two-module ($2 \times 8$B) configuration, performance peaks when $\lambda_1=0.3$, yielding an average accuracy of 72.3\%, while overly large or small values lead to suboptimal results. For the three-module ($3 \times 8$B) ensemble, the best performance of 73.6\% is obtained with $(\lambda_1=0.3, \lambda_2=0.3)$, suggesting that balanced weighting across modules is essential. These findings highlight that careful tuning of $\lambda$ is critical for maximizing the benefit of multi-model integration.

\subsection{Top-k Selection}
\begin{table*}[!t]
    \centering
    \caption{Performance comparison (\%) under different top-$k$ selections on logits.}
    \label{tab:logitsTopk}
    \small
    % \resizebox{0.6\textwidth}{!}{ % 宽度缩小一些更紧凑
    \begin{tabular}{l|l|lll}
        \toprule
        Params & k & SVAMP & MAWPS  & Avg. \\ 
        \midrule
        \multicolumn{5}{c}{\textbf{Qwen-2.5-7B as base model}} \\
        \midrule
        \multirow{8}{*}{$2 \times 7B$}
         & 1 & 87.9 & 92.9  & 90.4 \\ 
        ~ & 2 & 88.8 & 94.1  & \textbf{91.2} \\ 
        ~ & 3 & 88.3 & 93.3  & 90.8 \\ 
        ~ & 4 & 87.3 & 93.4  & 90.4 \\
        ~ & 5 & 87.4 & 93.3  & 90.4 \\
        ~ & 6 & 87.1 & 93.3  & 90.2 \\
        ~ & 7 & 87.2 & 93.3  & 90.3 \\
        ~ & 8 & 87.2 & 93.3  & 90.3 \\
        \bottomrule
    \end{tabular}
    % }
    \vspace{-10pt}
\end{table*}

% To analyze how $k$ influences the top-k logits fusion strategy—which selects high-confidence tokens for fusion—we conduct a sensitivity analysis using Qwen-2.5-7B as the base model in a 2$\times$7B ensemble. We test $k$ values from 1 to 8, fix other hyperparameters, and evaluate performance on the SVAMP and MAWPS mathematical reasoning datasets.

% Results confirm $k=2$ delivers the best performance, achieving higher accuracy on both SVAMP and MAWPS than other $k$ values. Specifically, 
% $k=1$ retains too few tokens and loses valuable contextual information, while $k\geq3$ leads to reduced or plateaued performance—this is because more tokens introduce low-confidence outputs that dilute the contribution of high-quality predictions.
We further examine how different top-$k$ settings affect decoding performance, with results summarized in Table~\ref{tab:logitsTopk}. 
Performance remains largely stable across $k$ values, showing only minor variations. 
The best average accuracy \textbf{91.2\%} is achieved at $k=2$, slightly outperforming other choices. 
This suggests that a small candidate set is sufficient to strike a balance between exploration and output stability during decoding. 
Notably, increasing $k$ beyond 3 yields diminishing returns: performance plateaus or even dips slightly, indicating that larger candidate pools offer little added value for reasoning tasks like SVAMP and MAWPS.

\subsection{Effect of Cross-Model Connection Sparsity}
\begin{table*}[!t]
    \centering
    \caption{Performance comparison (Performance comparison (\%) under different ) under different \textit{sparsity of cross-model connections sparsity of cross-model connections } settings, i.e., varying the interval of inter-layer interactions. settings, i.e., varying the interval of inter-layer interactions.}
    \label{tab:layerSkip}
    \small
    % \resizebox{0.6\textwidth}{!}{ % 宽度缩小一些更紧凑
    \begin{tabular}{l|l|lll}
        \toprule
        Params & $\phi$ & SVAMP & MAWPS  & Avg. \\ 
        \midrule
        \multicolumn{5}{c}{\textbf{Qwen-2.5-7B as base model}} \\
        \midrule
        \multirow{4}{*}{$2 \times 7B$}
         & 1 & 88.4 & 93.3 & 90.9 \\ 
         & 2 & 88.8 & 94.1 & 91.5 \\ 
         & 3 & 88.1 & 93.3 & 90.7 \\ 
         & 4 & 86.8 & 92.9 & 89.9 \\
        \bottomrule
    \end{tabular}
    % }
    \vspace{-10pt}
\end{table*}

We conduct experiments to test how the sparsity of cross-model connections controlled by $\phi$ affects performance, using Qwen-2.5-7B as the base model in a 2$\times$7B ensemble. We test $\phi=1$
 to $4$, fix other hyperparameters and evaluate on SVAMP and MAWPS.
Results show $\phi=2$ achieves the best performance, with higher accuracy on both datasets than other $\phi$ values. 
$\phi=1$ frequent cross-model fusion causes inefficient computation without extra gains, while $\phi\geq3$ over-sparse fusion loses valuable intermediate information, leading to reduced performance.

\subsection{$\alpha \& \beta$ Selection}
\begin{table*}[!t]
    \centering
    \caption{Performance comparison (\%) under different $\alpha$ selections.}
    \small
    \label{tab:alpha}
    \begin{tabular}{l|c|ccc}
        \toprule
        Params & $\alpha$ & SVAMP & MAWPS & Avg. \\
        \midrule
        \multicolumn{5}{c}{\textbf{Qwen-2.5-7B as base model}} \\
        \midrule
        \multirow{5}{*}{$2 \times 7$B}
         & 1.00 & 87.0 & 93.7 & 90.4 \\ 
         & 0.99 & 87.0 & 93.7 & 90.4 \\
         & 0.95 & 88.2 & 93.7 & 91.1 \\
         & 0.90 & \textbf{88.8} & \textbf{94.1} & \textbf{91.5} \\
         & 0.80 & 87.9 & 92.9 & 90.4 \\
         & 0.50 & 87.9 & 92.4 & 90.2 \\
        \bottomrule
    \end{tabular}
    \vspace{-10pt}
\end{table*}

\begin{table*}[!t]
    \centering
    \caption{Performance comparison (\%) under different $\beta$ selections.}
    \small
    \label{tab:beta}
    \begin{tabular}{l|c|ccc}
        \toprule
        Params & $\beta$ & SVAMP & MAWPS & Avg. \\
        \midrule
        \multicolumn{5}{c}{\textbf{Qwen-2.5-7B as base model}} \\
        \midrule
        \multirow{5}{*}{$2 \times 7$B}
         & 0.01 & 87.0 & 92.4 & 89.7 \\
         & 0.05 & 88.1 & 93.7 & 90.9 \\
         & 0.10 & 88.8 & 92.4 & \textbf{91.5} \\
         & 0.20 & 89.0 & 92.4 & 90.7 \\
         & 0.50 & 88.5 & 92.4 & 90.5 \\
        \bottomrule
    \end{tabular}
    \vspace{-10pt}
\end{table*}

In the loss function, the choices of $\alpha$ and $\beta$ form an $n^2$ search space, which makes exhaustive exploration computationally infeasible. To reduce the search cost, we first fix $\alpha=1.0$ and sweep across different $\beta$ values. After identifying the best-performing $\beta$, we then fix this value and sweep $\alpha$. The results summarized in Table~\ref{tab:alpha} and Table~\ref{tab:beta} show that \modelname achieves its best overall performance when both hyperparameters are set to $\alpha = 0.90$ and $\beta = 0.10$. These observations verify that our two-stage search strategy is effective and that moderate values of $\alpha$ and $\beta$ provide a good balance between suppression strength and optimization stability.

\subsection{Detailed Efficiency Results}

In this appendix, we provide a detailed comparison between single-model baselines (Qwen2.5-14B, Qwen2.5-7B), traditional ensemble methods (Vote, TCopilot, Unite), and our proposed \modelname under different execution strategies. Here, Seq denotes sequential execution, where multiple 7B sub-models are evaluated one after another on the same computation path; Par denotes parallel execution, where sub-models are evaluated concurrently; and GPU=1 indicates that all sub-models are colocated on a single GPU, whereas GPU $>$ 1 distributes different sub-models across multiple GPUs.

% 需在导言区加载 booktabs 宏包（\usepackage{booktabs}）
\begin{table*}[!t]
    \centering
    \caption{Efficiency Comparison of Different Models (Time/Memory Metrics)}
    \small
    \label{tab:efficiency}
    \setlength{\tabcolsep}{4pt} % 缩小列间距，平衡紧凑性
    \begin{tabular}{lcccc}
        \toprule
        Model & \makecell{End-to-End\\(s)} & \makecell{Token Cost\\(s)} & \makecell{Peak Train Mem\\(\%)} & \makecell{Peak Infer Mem\\(\%)} \\
        \midrule
        Qwen14B                             & 17.40 & 0.098 & 56.74\% & 36.97\% \\
        LLMBoost 7B+7B (near-par, 2 GPUs)   & 13.75 & 0.075 & 33.18\% & 20.72\% \\
        LLMBoost 7B+7B (near-par, 1 GPU)    & 14.87 & 0.081 & 54.11\% & 39.75\% \\
        LLMBoost 7B+7B+7B (near-par, 3 GPUs)& 18.88 & 0.103 & 33.18\% & 20.13\% \\
        LLMBoost 7B+7B+7B (near-par, 1 GPU) & 21.67 & 0.118 & 73.42\% & 58.48\% \\
        LLMBoost 7B+7B (seq)                & 22.53 & 0.125 & 60.29\% & 38.72\% \\
        VOTE (7B+7B)                        & 22.61 & 0.127 & 53.39\% & 39.14\% \\
        T-Copilot (7B+7B)                   & 17.21 & 0.097 & 64.41\% & 42.05\% \\
        UNITE (7B+7B)                       & 22.66 & 0.128 & 53.39\% & 39.12\% \\
        \bottomrule
    \end{tabular}
    \vspace{-10pt}
\end{table*}

In terms of end-to-end latency, Qwen2.5-14B serves as the single-model baseline (17.40 s, 0.098 s/token). Notably, \modelname(Qwen2.5-7B$\times$2 Par, GPU=2) achieves 13.75 s end-to-end latency and 0.075 s/token, delivering a two-model ensemble that is not only faster than Qwen2.5-14B but also significantly more efficient than traditional ensembles such as Vote (22.61 s, 0.127 s/token). This demonstrates that under genuine multi-GPU parallelism, the Par strategy can reap ensemble gains with manageable latency overhead while outperforming the single 14B model. By contrast, \modelname(Qwen2.5-7B$\times$2 Seq), which performs linear execution on a single device, increases end-to-end latency to 22.53 s—closely matching the behavior of Vote and Unite in the 22–23 s range—and exhibits the expected near-linear growth of runtime with respect to the number of sub-models. Similarly, \modelname(Qwen2.5-7B$\times$3 Par, GPU=1) reaches 21.67 s, illustrating that when all models reside on a single GPU, “parallel” execution is effectively limited by the same hardware budget and cannot realize true multi-stream speedup.

From the memory perspective, Qwen2.5-14B (the single-model baseline) requires 56.74\% peak training memory and 36.97\% peak inference memory. Under multi-GPU parallel settings, \modelname-7B$\times$2 (Par, GPU=2) and \modelname(Qwen2.5-(Qwen2.5-7B$\times$3 Par, GPU=3) both maintain training peak memory around 33.18\%, with inference peaks of 20.72\% and 20.13\%, respectively—far lower than the single Qwen2.5-14BQwen2.5-14B baseline. This indicates that distributing sub-models across GPUs allows Par execution to “spread” the memory footprint rather than stacking it on a single device. In contrast, single-GPU multi-model configurations (e.g., \modelname(Qwen2.5-(Qwen2.5-7B$\times$3 Par, GPU=1) with 73.42\% training and 58.48\% inference peak memory, or \modelname(Qwen2.5-(Qwen2.5-7B$\times$2 Seq) with 60.29\% training) and traditional ensembles (Vote at 53.39\%, T-Copilot at 64.41\% training peak) exhibit substantial cumulative memory overhead as the number of models increases.

Overall, these observations support three key conclusions: (1) Seq (sequential) execution on a single GPU leads to almost linear growth in both latency and memory usage with respect to the ensemble size; (2) Par (parallel) execution over multiple GPUs achieves faster latency than the single Qwen2.5--14B model while keeping per-device memory consumption significantly lower; and (3) under the same hardware budget, \modelname strikes a strictly better trade-off between end-to-end latency, per-token efficiency, and memory consumption than Qwen2.5--14B and existing ensemble baselines, making it a more practical option for real-world multi-GPU deployment.

\begin{table}[H]
    \caption{Accuracy (\%) across different ensemble sizes on the arithmetic reasoning dataset.}
    \label{tab:numberOfEnsembleModels}
    \centering
    \small  
    \begin{tabular}{l|c|cccc|c}
        \toprule
        Model & Params & AQuA & GSM8K & MAWPS & SVAMP & Avg. \\
        \midrule
        \multicolumn{7}{c}{\textbf{Llama-3.1-8B}} \\
        \midrule
        \modelname & $1\times 8\mathrm{B}$ & 42.3 & 63.7 & 89.5 & 77.4 & 68.2 \\
        \modelname & $2\times 8\mathrm{B}$ & 46.8 & 68.8 & 90.1 & 80.1 & 71.5 \\
        \modelname & $3\times 8\mathrm{B}$ & 47.6 & 68.5 & 92.0 & 81.3 & 72.4 \\
        \modelname & $4\times 8\mathrm{B}$ & 47.1 & 69.3 & 92.1 & 80.9 & 72.4 \\
        \midrule
        \multicolumn{7}{c}{\textbf{Qwen2.5-7B}} \\
        \midrule
        \modelname & $1\times 7\mathrm{B}$ & 56.8 & 75.1 & 92.0 & 86.1 & 77.5 \\
        \modelname & $2\times 7\mathrm{B}$ & 59.8 & 78.1 & 94.1 & 88.8 & 80.2 \\
        \modelname & $3\times 7\mathrm{B}$ & 61.8 & 78.9 & 93.2 & 88.4 & 80.6 \\
        \modelname & $4\times 7\mathrm{B}$ & 62.2 & 78.7 & 93.5 & 88.5 & 80.7 \\
        \bottomrule
    \end{tabular}%
    
    \vspace{-10pt}
\end{table}

\subsection{Study on Number of LLMs.}
We further explore the impact of varying the number of sub-models $n$ in \textsc{\modelname}.
As shown in Table~\ref{tab:numberOfEnsembleModels}, performance improves consistently when increasing $n$ from 1 to 4.
This monotonic trend confirms that subsequent models are able to refine and strengthen the outputs of their predecessors.
However, the improvement exhibits diminishing returns beyond a certain number of sub-models ($n \geq 3$), implying that most of the corrective and reinforcing capacity is already realized in the earlier stages of the ensemble.
In addition, increasing $n$ inevitably leads to higher computational and latency costs.
By experimental results, we observe that $n=3$ achieves an effective balance between accuracy and efficiency, yielding strong gains without excessive overhead.
Overall, these findings demonstrate that \textsc{\modelname} can achieve substantial benefits even with a relatively small ensemble size.

\subsection{Detailed Description of e2e testing}

The end-to-end toolchain scheduling tests, built on \dianxinlab’s dedicated test sets, aim to evaluate the capability of large language model to accurately execute tool-call tasks in \dianxinlab’s internal workflows.

The test follows a standardized process: a user query and an expected tool call are input into the model via OpenAI’s /v1/chat/completions interface, and the returned tool calls are then compared with the expected ones to determine whether a test case passes.

Each test input includes a user query and an expected\_tool\_call, while the output is the model’s response, containing the actual tool\_calls and a finish reason.

When using two Qwen-2.5-7B models, the overall pass rate of 94 test cases reaches 24\% (23/94); when using one Qwen-2.5-7B model, the pass rate drops to 19\% (18/94)—demonstrating that multi-model setups outperform single-model setups in tool-call accuracy.

\newpage

\section{Proofs of the Main Theoretical Results}
\label{app:proof_of_thm_and_cor}
\subsection{Formal Setup for Proofs}
This section provides detailed proofs for the theorems presented in the main body. The analysis is conducted for an arbitrary decoding step $t$ during inference. For notational brevity, we adopt the following conventions throughout the proofs.

For the analysis at a given decoding step $t$, let $m_i$ denote an individual model and $A_i$ its sampling space. The logits produced by this model at step $t$ are given by $z^{(i)}_t$. We consider an ensemble of models $\mathcal{M}_i = \{m_0, \dots, m_i\}$, with a corresponding sampling space $\mathcal{A}_i$. The accumulated logits for this ensemble, $\hat{z}^{(i)}_t$, are defined recursively as $\hat{z}^{(i)}_t = \hat{z}^{(i-1)}_t + \lambda_i z^{(i)}_t$, starting with the base case $\hat{z}^{(0)}_t = z^{(0)}_t$. From these logits, the ensemble's predicted probability distribution is calculated as $\hat{P}^{(i)}_t = \mathrm{softmax}(\hat{z}^{(i)}_t)$. Let $\hat{y}^{(i-1)}_{<t}$ be the sequence of tokens generated by the ensemble $\mathcal{M}_{i-1}$ up to step $t-1$. The ground truth probability distribution for the token at step $t$, conditioned on an input $X_\tau$ and the generated prefix, is $p^*_t(\cdot \mid X_\tau, \hat{y}^{(i-1)}_{<t})$, which we denote by the one-hot vector $p^*_{t}$. The residual error of the ensemble $\mathcal{M}_{i-1}$ can thus be expressed as the difference between the true and the predicted distributions, $e^{(i-1)}_t = p^*_{t} - \hat{P}^{(i-1)}_t$. To quantify how a new model $m_i$ influences the ensemble's prediction, we define its effective linear contribution to the probability space as $g^{(i)}_t \triangleq J_{\mathrm{softmax}}(\hat{z}^{(i-1)}_t) z^{(i)}_t$, where $J_{\mathrm{softmax}}(\hat{z}^{(i-1)}_t)$ is the Jacobian matrix of the softmax function evaluated at the accumulated logits of the preceding ensemble, $\hat{z}^{(i-1)}_t$.

The expectation $\mathbb{E}[\cdot]$ is taken over all relevant sources of randomness (data distribution $\mathcal{D}$ and model training distributions $\mathcal{A}_i$). We omit the explicit conditioning on the input $X_\tau$ and the generated prefix $\hat{y}^{(i-1)}_{<t}$ for clarity, though all probabilistic quantities are implicitly dependent on this context.

\subsection{Preliminary Lemma for MSE Change}

\begin{lemma}[Asymptotic Form of Single-Dimension MSE Change]
\label{lemma:mse_change}
The change in MSE for a single vocabulary dimension $v$ at step $t$, denoted $\Delta_{\mathrm{MSE}, t, v} = \mathbb{E}[(e^{(i)}_{t,v})^2] - \mathbb{E}[(e^{(i-1)}_{t,v})^2]$, resulting from the addition of successor $m_i$ with a small scaling coefficient $\lambda_{i}$, can be expressed as:
\begin{equation}
\Delta_{\mathrm{MSE}, t, v} = -2\lambda_{i} \mathbb{E}[e^{(i-1)}_{t,v} g^{(i)}_{t,v}] + O(\lambda_{i}^2).
\end{equation}
\end{lemma}
\begin{proof}
The new ensemble probability is $\hat{P}^{(i)}_t = \mathrm{softmax}(\hat{z}^{(i-1)}_t + \lambda_{i} z^{(i)}_t)$. Using a first-order Taylor expansion, the new residual vector $e^{(i)}_t = p^*_t - \hat{P}^{(i)}_t$ can be written as:
\begin{align}
e^{(i)}_t &= p^*_t - \left( \hat{P}^{(i-1)}_t + \lambda_{i} J_{\mathrm{softmax}}(\hat{z}^{(i-1)}_t) z^{(i)}_t + R(\lambda_{i} z^{(i)}_t) \right) \\
&= (p^*_t - \hat{P}^{(i-1)}_t) - \lambda_{i} g^{(i)}_t - R(\lambda_{i} z^{(i)}_t) \\
&= e^{(i-1)}_t - \lambda_{i} g^{(i)}_t - R(\lambda_{i} z^{(i)}_t),
\end{align}
where the remainder vector $R(\cdot)$ satisfies $\|R(\Delta z)\|_2 = O(\|\Delta z\|_2^2)$. Let $R_{t,v}$ be the $v$-th component of $R(\lambda_i z^{(i)}_t)$. The change in MSE is $\Delta_{\mathrm{MSE}, t, v} = \mathbb{E}[(e^{(i-1)}_{t,v} - \lambda_{i}g^{(i)}_{t,v} - R_{t,v})^2] - \mathbb{E}[(e^{(i-1)}_{t,v})^2]$. Expanding the squared term and using linearity of expectation gives:
\begin{equation}
\Delta_{\mathrm{MSE}, t, v} = -2\lambda_i \mathbb{E}[e^{(i-1)}_{t,v} g^{(i)}_{t,v}] + \lambda_i^2 \mathbb{E}[(g^{(i)}_{t,v})^2] - 2\mathbb{E}[e^{(i-1)}_{t,v} R_{t,v}] + 2\lambda_i \mathbb{E}[g^{(i)}_{t,v} R_{t,v}] + \mathbb{E}[R_{t,v}^2].
\end{equation}
As shown in Lemma \ref{app:lemma_expectation_orders}, all terms involving the remainder $R_{t,v}$ are of order $O(\lambda_i^2)$ or higher. For a sufficiently small $\lambda_{i} > 0$, the linear term dominates, leading to the stated asymptotic form.
\end{proof}

\subsection{Proof of Theorem \ref{thm:mse_reduction}}

\begin{theorem}[Restate]
\label{thm:mse_reduction_restate}
Under Assumption \ref{ass:correction_efficacy}, for each stage $i \geq 1$, decoding step $t$, and vocabulary dimension $v$, there exists an upper bound $\lambda_{i, t, v}^* > 0$ for the scaling coefficient. For any weight $\lambda_{i} \in (0, \lambda_{i, t, v}^*)$, the MSE of the new ensemble $\mathcal{M}_{i}$ is strictly lower than that of $\mathcal{M}_{i-1}$ for that dimension.
\end{theorem}
\begin{proof}
We analyze the change in MSE, $\Delta_{\mathrm{MSE}, t, v} = \mathbb{E}[(e^{(i)}_{t,v})^2] - \mathbb{E}[(e^{(i-1)}_{t,v})^2]$. Our goal is to find a $\lambda_{i}>0$ such that $\Delta_{\mathrm{MSE}, t, v} < 0$. By Lemma \ref{lemma:mse_change}, the asymptotic form is:
\begin{equation}
\Delta_{\mathrm{MSE}, t, v} = -2\lambda_{i} \mathbb{E}[e^{(i-1)}_{t,v}g^{(i)}_{t,v}] + O(\lambda_{i}^2).
\end{equation}
For a sufficiently small $\lambda_{i}>0$, the sign of $\Delta_{\mathrm{MSE}, t, v}$ is determined by $-\mathbb{E}[e^{(i-1)}_{t,v}g^{(i)}_{t,v}]$. We now show this expectation is strictly positive.

By the law of total expectation, we can first condition on the parameters $\Theta_{i-1}$ and the generated context $(X_\tau, \hat{y}^{(i-1)}_{<t})$, which determine the value of the error $e^{(i-1)}_{t,v}$.

\begin{align}
\mathbb{E}[e^{(i-1)}_{t,v}g^{(i)}_{t,v}] &= \mathbb{E}\left[ \mathbb{E}[e^{(i-1)}_{t,v}g^{(i)}_{t,v} \mid \Theta_{i-1}, X_\tau, \hat{y}^{(i-1)}_{<t}] \right] \\
&= \mathbb{E}\left[ e^{(i-1)}_{t,v} \cdot \mathbb{E}[g^{(i)}_{t,v} \mid \Theta_{i-1}, X_\tau, \hat{y}^{(i-1)}_{<t}] \right]  \label{eq:law_of_total_exp}
\end{align}
In \eqref{eq:law_of_total_exp}, $e^{(i-1)}_{t,v}$ is treated as a constant inside the inner expectation because it is fully determined by the conditioning variables. Let's denote the inner expectation of the contribution term as $\bar{g}^{(i)}_{t,v} \triangleq \mathbb{E}[g^{(i)}_{t,v} \mid \Theta_{i-1}, X_\tau, \hat{y}^{(i-1)}_{<t}]$. The expression simplifies to $\mathbb{E}[e^{(i-1)}_{t,v} \bar{g}^{(i)}_{t,v}]$. We proceed as follows:$$
\begin{aligned}
\mathbb{E}[e^{(i-1)}_{t,v} \bar{g}^{(i)}_{t,v}] &= \mathbb{E}[e^{(i-1)}_{t,v} (e^{(i-1)}_{t,v} - (e^{(i-1)}_{t,v} - \bar{g}^{(i)}_{t,v}))] \\
&= \mathbb{E}[(e^{(i-1)}_{t,v})^2] - \mathbb{E}[e^{(i-1)}_{t,v}(e^{(i-1)}_{t,v} - \bar{g}^{(i)}_{t,v})]
\end{aligned}
$$Applying the Cauchy-Schwarz inequality to the second term gives:$$
\left| \mathbb{E}[e^{(i-1)}_{t,v}(e^{(i-1)}_{t,v} - \bar{g}^{(i)}_{t,v})] \right| \le \sqrt{\mathbb{E}[(e^{(i-1)}_{t,v})^2] \cdot \mathbb{E}[(e^{(i-1)}_{t,v} - \bar{g}^{(i)}_{t,v})^2]}
$$From the definitions in Assumption \ref{ass:correction_efficacy}, we recognize that $\mathbb{E}[(e^{(i-1)}_{t,v})^2] = \epsilon_{i-1, t, v}^2 + \sigma_{i-1, t, v}^2$, and the bias term is precisely $\epsilon_{g^{(i)}_{t, v}}^2 = \mathbb{E}[(e^{(i-1)}_{t,v} - \bar{g}^{(i)}_{t,v})^2]$. This establishes a rigorous lower bound for our term of interest:
\begin{align}
\mathbb{E}[e^{(i-1)}_{t,v}g^{(i)}_{t,v}] &\ge \mathbb{E}[(e^{(i-1)}_{t,v})^2] - \sqrt{\mathbb{E}[(e^{(i-1)}_{t,v})^2] \cdot \epsilon_{g^{(i)}_{t, v}}^2} \\
&= \sqrt{\mathbb{E}[(e^{(i-1)}_{t,v})^2]} \left( \sqrt{\mathbb{E}[(e^{(i-1)}_{t,v})^2]} - \epsilon_{g^{(i)}_{t, v}} \right) \\
&= \sqrt{\epsilon_{i-1, t, v}^2 + \sigma_{i-1, t, v}^2} \left( \sqrt{\epsilon_{i-1, t, v}^2 + \sigma_{i-1, t, v}^2} - \epsilon_{g^{(i)}_{t, v}} \right)
\end{align}

Assumption \ref{ass:correction_efficacy} states that $\epsilon_{g^{(i)}_{t, v}} < \sqrt{\epsilon_{i-1, t, v}^2 + \sigma_{i-1, t, v}^2}$, which guarantees that the term in the parentheses is strictly positive. Therefore, $\mathbb{E}[e^{(i-1)}_{t,v}g^{(i)}_{t,v}]$ is strictly positive.
As $\mathbb{E}[e^{(i-1)}_{t,v}g^{(i)}_{t,v}] > 0$, the change in MSE, $\Delta_{\mathrm{MSE}, t, v}$, will be negative for any $\lambda_{i}$ in an interval $(0, \lambda_{i,t,v}^*)$ for some positive constant $\lambda_{i,t,v}^*$. This proves the theorem.
\end{proof}

\subsection{Proof of Corollary \ref{cor:uniform_lambda_restate}}

\begin{corollary}[Restate]
\label{cor:uniform_lambda_restate}
For each stage $i \geq 1$ and decoding step $t$, there exists a single, per-token scalar $\lambda_{i,t}^* > 0$, such that for any $\lambda_{i} \in (0, \lambda_{i,t}^*)$, the total MSE of the new ensemble $\mathcal{M}_{i}$ is strictly less than that of the predecessor ensemble $\mathcal{M}_{i-1}$.
\end{corollary}
\begin{proof}
From Theorem \ref{thm:mse_reduction}, for each dimension $v \in \{1, \dots, |\mathcal{V}|\}$, there exists an upper bound $\lambda_{i, t, v}^* > 0$. We define a per-token upper bound $\lambda_{i,t}^*$ as the minimum of these per-dimension bounds:
\begin{equation}
  \lambda_{i,t}^* \triangleq \min_{v=1, \dots, |\mathcal{V}|} \{ \lambda_{i, t, v}^* \}.  
\end{equation}
Since this is the minimum over a finite set of strictly positive numbers, $\lambda_{i,t}^*$ is itself strictly positive.

For any scaling coefficient $\lambda_{i}$ such that $0 < \lambda_{i} < \lambda_{i,t}^*$, the condition $\lambda_{i} < \lambda_{i, t, v}^*$ holds for all dimensions $v$. By Theorem \ref{thm:mse_reduction}, this guarantees that the change in MSE for each dimension is negative:
\begin{equation}
    \Delta_{\mathrm{MSE}, t, v} < 0 \quad \text{for all } v \in \{1, \dots, |\mathcal{V}|\}.
\end{equation}
The total change in MSE at step $t$ is the sum of these per-dimension changes:
\begin{equation}
    \Delta_{\mathrm{MSE}, t} = \mathrm{MSE}(\mathcal{M}_{i};t) - \mathrm{MSE}(\mathcal{M}_{i-1};t) = \sum_{v=1}^{|\mathcal{V}|} \Delta_{\mathrm{MSE}, t, v}.
\end{equation}
As the sum of strictly negative terms, the total change $\Delta_{\mathrm{MSE}, t}$ must be strictly negative. Therefore, for any $\lambda_{i} \in (0, \lambda_{i,t}^*)$, we have $\mathrm{MSE}(\mathcal{M}_{i};t) < \mathrm{MSE}(\mathcal{M}_{i-1};t)$. This concludes the proof.
\end{proof}

% \newpage
\subsection{Technical Lemmas and Bounds for the Proofs} % (将原 B 部分作为子节)
\label{app:technical_details}
This appendix provides rigorous justifications for the technical claims made in the proof of Lemma \ref{lemma:mse_change}.

\subsubsection{Quadratic Bound on the Taylor Remainder of Softmax}
\label{app:sub_quadratic_bound}

\begin{definition}[Hessian as a Symmetric Bilinear Form]
\label{def:hessian_bilinear}
Let $f: \mathbb{R}^n \to \mathbb{R}^m$ be a twice continuously differentiable function ($C^2$). The second derivative of $f$ at a point $u \in \mathbb{R}^n$, denoted as $D^2 f(u)$, is a bilinear form that maps two vectors from $\mathbb{R}^n$ to a vector in $\mathbb{R}^m$.

For any two vectors $v, w \in \mathbb{R}^n$, the resulting vector $D^2 f(u)[v, w] \in \mathbb{R}^m$ is defined component-wise:
$$\left( D^2 f(u)[v, w] \right)_i = \sum_{j=1}^{n} \sum_{k=1}^{n} \frac{\partial^2 f_i(u)}{\partial u_j \partial u_k} v_j w_k, \quad \text{for } i=1, \dots, m.$$
Since $f$ is $C^2$, by Clairaut's theorem on the equality of mixed partials, this bilinear form is symmetric, i.e., $D^2 f(u)[v, w] = D^2 f(u)[w, v]$. In the context of the Taylor expansion remainder, this form is evaluated with identical vectors, as in $D^2 f(u)[\Delta z, \Delta z]$.
\end{definition}

\begin{definition}[Operator Norm of the Hessian Tensor]
\label{def:hessian_op_norm}
The operator norm of the second derivative $D^2 f(u)$, induced by the vector $\ell_2$-norm, is defined as:
$$\|D^2 f(u)\|_{\mathrm{op}} = \sup_{\|v\|_2=1, \|w\|_2=1} \|D^2 f(u)[v, w]\|_2.$$For any fixed $u$ in a finite-dimensional space, this norm is finite. The crucial property for the Taylor remainder bound is the \textbf{global boundedness} of this norm, i.e., the existence of a constant $L$ such that:
$$L \coloneqq \sup_{u \in \mathbb{R}^n} \|D^2 f(u)\|_{\mathrm{op}} < \infty.$$
\end{definition}

The global boundedness of the Hessian for $f(z) = \mathrm{softmax}(z)$ is established by observing that its components are polynomials of the outputs $p_i = \mathrm{softmax}(z)_i$. The first and second derivatives of the softmax function are given by:
\begin{align}
\frac{\partial p_i}{\partial z_j} &= p_i(\delta_{ij} - p_j) \\
\frac{\partial^2 p_i}{\partial z_j \partial z_k} &= p_i \big[ (\delta_{ik} - p_k)(\delta_{ij} - p_j) - p_j(\delta_{jk} - p_k) \big]
\end{align}
where $p = \mathrm{softmax}(z)$ and $\delta_{ij}$ is the Kronecker delta.

The components of the Hessian tensor are evidently polynomials in the components of $p$. The image of the softmax function is the open standard simplex. Since these polynomial expressions are continuous, they are bounded on the closure of this image, which is the standard simplex—a compact set. This ensures that each component of the Hessian tensor is globally bounded, which in turn guarantees that its operator norm is globally bounded by a constant $L < \infty$.

\begin{lemma}[Quadratic Bound on the Taylor Remainder of Softmax]
\label{app:lemma_quadratic_bound}
The Taylor remainder $R(\Delta z) = \mathrm{softmax}(z + \Delta z) - \mathrm{softmax}(z) - J_{\mathrm{softmax}}(z)\Delta z$ satisfies the quadratic bound $\|R(\Delta z)\|_2 \le C \|\Delta z\|_2^2$ for a global constant $C$.
\end{lemma}
\begin{proof}
Let $f(z) = \mathrm{softmax}(z)$. Since $f$ is infinitely differentiable ($C^\infty$), we can express the remainder term using Taylor's theorem with integral remainder:
$$
R(\Delta z) = \int_0^1 (1-t) D^2 f(z+t\Delta z)[\Delta z, \Delta z] dt,
$$
where $J_f(z)$ is the Jacobian matrix of $f$ at $z$, and $D^2 f(u)[\cdot, \cdot]$ is the second derivative (a bilinear form corresponding to the Hessian tensor) at a point $u$. By taking the vector $\ell_2$-norm, we get:
$$
\|R(\Delta z)\|_2 \le \left( \int_0^1 (1-t) dt \right) \cdot \sup_{t \in [0,1]} \|D^2 f(z+t\Delta z)\|_{\mathrm{op}} \cdot \|\Delta z\|_2^2 = \frac{1}{2} \sup_{u \in [z, z+\Delta z]} \|D^2 f(u)\|_{\mathrm{op}} \cdot \|\Delta z\|_2^2,
$$
where $\|\cdot\|_{\mathrm{op}}$ is the operator norm. The entries of the Jacobian and the Hessian of the softmax function are polynomials of its components, $p_i = \mathrm{softmax}(z)_i$. Since the output vector $p$ is always on the standard simplex (i.e., $p_i \in [0,1]$ and $\sum_i p_i = 1$), which is a compact set, the values of these derivatives are globally bounded for any input $z \in \mathbb{R}^{|\mathcal{V}|}$. Therefore, there exists a global constant $L = \sup_{u \in \mathbb{R}^{|\mathcal{V}|}} \|D^2 f(u)\|_{\mathrm{op}} < \infty$. This gives the desired quadratic bound with $C=L/2$, formally justifying that $\|R(\Delta z)\|_2 = O(\|\Delta z\|_2^2)$.
\end{proof}

\subsubsection{Component Bound from Vector Norm Bound}
\begin{lemma}[Component Bound from Vector Norm Bound]
\label{app:lemma_component_bound}
For any vector $v \in \mathbb{R}^{d}$, each component $v_m$ is bounded by its $\ell_2$-norm: $|v_m| \le \|v\|_2$. Consequently, if $\|R(\Delta z)\|_2 = O(\|\Delta z\|_2^2)$, then each component $R_m(\Delta z)$ is also $O(\|\Delta z\|_2^2)$.
\end{lemma}
\begin{proof}
The proof is a direct application of the definition of the $\ell_2$-norm: $\|v\|_2 = \sqrt{\sum_{i=1}^d v_i^2}$. Since all terms in the sum are non-negative, we have $\sqrt{v_m^2} \le \sqrt{\sum_{i=1}^d v_i^2}$, which simplifies to $|v_m| \le \|v\|_2$. Applying this to the remainder vector $R$, we get $|R_m| \le \|R\|_2$. From Lemma \ref{app:lemma_quadratic_bound}, there exists a constant $C$ such that $\|R(\Delta z)\|_2 \le C \|\Delta z\|_2^2$. Therefore, $|R_m(\Delta z)| \le C \|\Delta z\|_2^2$, which means $R_m = O(\|\Delta z\|_2^2)$.
\end{proof}

\subsection{Expectation Orders of Remainder Terms}
\begin{lemma}[Expectation Orders of Remainder Terms]
\label{app:lemma_expectation_orders}
Under the finite moment conditions in Assumption \ref{ass:correction_efficacy}, for any decoding step $t$ and vocabulary dimension $v$, the expectations of terms involving the Taylor remainder component $R_{t,v} = R_v(\lambda_i z^{(i)}_t)$ have the following orders of magnitude with respect to the scaling coefficient $\lambda_i$:
\begin{align}
    \mathbb{E}[e^{(i-1)}_{t,v}R_{t,v}] &= O(\lambda_i^2), \\
    \mathbb{E}[\lambda_i g^{(i)}_{t,v}R_{t,v}] &= O(\lambda_i^3), \\
    \mathbb{E}[R_{t,v}^2] &= O(\lambda_i^4).
\end{align}
\end{lemma}
\begin{proof}
We analyze each term by applying the Cauchy-Schwarz inequality along with bounds on the Taylor remainder and model outputs. Let the full remainder vector be $R_t = R(\lambda_i z^{(i)}_t)$.

\paragraph{Analysis of $\mathbb{E}[e^{(i-1)}_{t,v}R_{t,v}]$.}
By the Cauchy-Schwarz inequality, $|\mathbb{E}[e^{(i-1)}_{t,v}R_{t,v}]| \le \sqrt{\mathbb{E}[(e^{(i-1)}_{t,v})^2] \mathbb{E}[R_{t,v}^2]}$. The term $\mathbb{E}[(e^{(i-1)}_{t,v})^2]$ is the MSE of the predecessor ensemble for this dimension, which is finite ($O(1)$) by our setup. For the remainder term, we use Lemma \ref{app:lemma_component_bound} to state that $|R_{t,v}| \le \|R_t\|_2$. Subsequently, Lemma \ref{app:lemma_quadratic_bound} provides that $\|R_t\|_2 \le C \|\lambda_i z^{(i)}_t\|_2^2 = C \lambda_i^2 \|z^{(i)}_t\|_2^2$ for some global constant $C$. Squaring and taking the expectation, we get $\mathbb{E}[R_{t,v}^2] \le \mathbb{E}[\|R_t\|_2^2] \le C^2 \lambda_i^4 \mathbb{E}[\|z^{(i)}_t\|_2^4]$. By Assumption \ref{ass:correction_efficacy}, $\mathbb{E}[\|z^{(i)}_t\|_2^4]$ is finite. Thus, $\mathbb{E}[R_{t,v}^2] = O(\lambda_i^4)$. Substituting these orders back into the Cauchy-Schwarz inequality gives $|\mathbb{E}[e^{(i-1)}_{t,v}R_{t,v}]| \le \sqrt{O(1) \cdot O(\lambda_i^4)} = O(\lambda_i^2)$.

\paragraph{Analysis of $\mathbb{E}[\lambda_i g^{(i)}_{t,v}R_{t,v}]$.}
This expectation can be written as $\lambda_i \mathbb{E}[g^{(i)}_{t,v}R_{t,v}]$. Applying the Cauchy-Schwarz inequality, we have $|\lambda_i \mathbb{E}[g^{(i)}_{t,v}R_{t,v}]| \le \lambda_i \sqrt{\mathbb{E}[(g^{(i)}_{t,v})^2] \mathbb{E}[R_{t,v}^2]}$. From the analysis above, we know $\mathbb{E}[R_{t,v}^2] = O(\lambda_i^4)$. For the term involving $g^{(i)}_{t,v}$, we note that $g^{(i)}_t = J_{\mathrm{softmax}}(\hat{z}^{(i-1)}_t) z^{(i)}_t$. The operator norm of the softmax Jacobian is globally bounded by a constant, say $C_J$. Thus, $\|g^{(i)}_t\|_2 \le C_J \|z^{(i)}_t\|_2$. This implies $\mathbb{E}[(g^{(i)}_{t,v})^2] \le \mathbb{E}[\|g^{(i)}_t\|_2^2] \le C_J^2 \mathbb{E}[\|z^{(i)}_t\|_2^2]$. Since the fourth-order moment of $z^{(i)}_t$ is finite, its second-order moment is also finite, making $\mathbb{E}[(g^{(i)}_{t,v})^2]$ a finite constant ($O(1)$). Combining these results, the overall expression is of order $\lambda_i \sqrt{O(1) \cdot O(\lambda_i^4)} = \lambda_i \cdot O(\lambda_i^2) = O(\lambda_i^3)$.

\paragraph{Analysis of $\mathbb{E}[R_{t,v}^2]$.}
As established in the first paragraph of this proof, the bound $|R_{t,v}| \le C \lambda_i^2 \|z^{(i)}_t\|_2^2$ leads directly to the expectation bound $\mathbb{E}[R_{t,v}^2] \le C^2 \lambda_i^4 \mathbb{E}[\|z^{(i)}_t\|_2^4]$. Given the finite fourth-order moment assumption on the logits $z^{(i)}_t$, we confirm that $\mathbb{E}[R_{t,v}^2] = O(\lambda_i^4)$.
\end{proof}

\newpage

\section{Justification for the Core Assumption: An Optimization Perspective}

The theoretical analysis in the main body (Section \ref{sec:analysis}) is conducted in the inference setting, establishing performance guarantees for the final trained ensemble. That analysis relies on Assumption~\ref{ass:correction_efficacy}, which posits that each successor model is an effective error corrector. In this section, we provide a supporting analysis from an optimization perspective during the training phase. We demonstrate that our proposed training objective (Eq.~\ref{equ:totalloss}) naturally steers the model parameters in a direction that encourages the conditions of Assumption~\ref{ass:correction_efficacy} to be met. While this analysis is conducted under a teacher-forcing training setting (using ground-truth prefixes), its conclusions provide a strong heuristic justification for the validity of our core assumption in the auto-regressive inference setting.

\label{app:support}

\subsection{Theoretical Statements for the Training Setting}
\label{subsec:analysis_for_training}

This subsection presents the formal statements of the assumption, theorem, and corollary for the training-time setting, which relies on the Teacher Forcing paradigm. The proofs follow the same structure as those for the inference-time analysis presented in the preceding appendix.

For the analysis at a given decoding step $t$, the context is the ground-truth prefix $(X_\tau, y^*_{<t})$. The ground-truth target is the one-hot vector $y^*_t$. The residual error of the ensemble $\mathcal{M}_{i-1}$ is defined as $e^{(i-1)}_t = y^*_{t} - \hat{P}^{(i-1)}_t$, where $\hat{P}^{(i-1)}_t$ is the probability distribution of the ensemble.

\begin{assumption}
\label{ass:correction_efficacy_training}
For any stage $i \geq 1$, any decoding step $t$, and any vocabulary dimension $v \in \{1, \dots, |\mathcal{V}|\}$, we assume that the successor model $m_{i}$, when trained, produces an effective linear contribution $g^{(i)}_t$ that is a valid corrective term of the residual error vector $e^{(i-1)}_t$. Specifically, we assume its systematic bias in fitting this residual is less than the total uncertainty of the residual. Formally, given an input context $(X_\tau, y^*_{<t})$:
\begin{equation}
\epsilon_{g^{(i)}_{t, v}} < \sqrt{\epsilon_{i-1, t, v}^2 + \sigma_{i-1, t, v}^2},
\end{equation}
where the bias ($\epsilon^2$) and variance ($\sigma^2$) terms are defined as:
\begin{align}
% Bias of the preceding ensemble M_{i-1}
\epsilon_{i-1, t, v}^2 &:= \mathbb{E}_{(X_\tau,Y_\tau)\sim\mathcal{D}} \left[ \left( y^*_{t,v} - \mathbb{E}_{\Theta_{i-1} \sim \mathcal{A}_{i-1}}[\hat{P}^{(i-1)}_{t,v} \mid X_\tau, y^*_{<t}] \right)^2 \right] < \infty, \\
% Variance of the preceding ensemble M_{i-1}
\sigma_{i-1, t, v}^2 &:= \mathbb{E}_{(X_\tau,Y_\tau)\sim\mathcal{D}} \left[ \mathrm{Var}_{\Theta_{i-1} \sim \mathcal{A}_{i-1}}[\hat{P}^{(i-1)}_{t,v} \mid X_\tau, y^*_{<t}] \right] < \infty , \\
% Bias of the new model m_i's contribution h^i_t in fitting the residual
\epsilon_{g^{(i)}_{t, v}}^2 &:= \mathbb{E}_{\substack{\Theta_{i-1} \sim \mathcal{A}_{i-1} \\ (X_\tau,Y_\tau)\sim\mathcal{D}}} \left[ \left( e^{(i-1)}_{t,v} - \mathbb{E}_{\theta_i \sim A_i}[g^{(i)}_{t,v} \mid \Theta_{i-1}, X_\tau, y^*_{<t}] \right)^2 \right] < \infty.
\end{align}
Furthermore, we assume that the logits generated by the successor model have a finite fourth-order moment, i.e., $\mathbb{E}[\|z^{(i)}_t(\cdot | X_\tau, y^*_{<t})\|_2^4] < \infty$.
\end{assumption}

\begin{theorem}
\label{thm:mse_reduction_training}
Under Assumption \ref{ass:correction_efficacy_training}, for each stage $i \geq 1$, decoding step $t$, and vocabulary dimension $v$, there exists an upper bound $\lambda_{i, t, v}^* > 0$. For any scaling coefficient $\lambda_{i} \in (0, \lambda_{i, t, v}^*)$, the expected squared error (MSE) of the new ensemble $\mathcal{M}_{i}$ is strictly lower than that of the predecessor ensemble $\mathcal{M}_{i-1}$ for that dimension. Specifically:
\begin{equation}
\begin{aligned}
\mathbb{E}_{\substack{\Theta_{i-1} \sim \mathcal{A}_{i-1} \\ \theta_i \sim A_i \\ (X_\tau,Y_\tau)\sim\mathcal{D}}}
\left[ \left( y^*_{t,v} - \hat{P}^{(i)}_{t,v} \right)^2 \mid X_\tau, y^*_{<t} \right]
<
\mathbb{E}_{\substack{\Theta_{i-1} \sim \mathcal{A}_{i-1} \\ (X_\tau,Y_\tau)\sim\mathcal{D}}}
\left[ \left( y^*_{t,v} - \hat{P}^{(i-1)}_{t,v} \right)^2 \mid X_\tau, y^*_{<t} \right].
\end{aligned}
\end{equation}
\end{theorem}

\begin{corollary}
\label{cor:uniform_lambda_training}
For each stage $i \geq 1$ and decoding step $t$, there exists a single, per-token scalar $\lambda_{i,t}^* > 0$, independent of vocabulary dimension. For any scaling coefficient $\lambda_{i} \in (0, \lambda_{i,t}^*)$, the total MSE of the new ensemble $\mathcal{M}_{i}$ for that token prediction is strictly less than that of the predecessor ensemble $\mathcal{M}_{i-1}$:
\begin{equation}
\begin{aligned}
\mathbb{E}_{\substack{\Theta_{i-1} \sim \mathcal{A}_{i-1} \\ \theta_i \sim A_i \\ (X_\tau,Y_\tau)\sim\mathcal{D}}}
\left[ \| y^*_t - \hat{P}^{(i)}_{t} \|_2^2 \mid X_\tau, y^*_{<t} \right]
<
\mathbb{E}_{\substack{\Theta_{i-1} \sim \mathcal{A}_{i-1} \\ (X_\tau,Y_\tau)\sim\mathcal{D}}}
\left[ \| y^*_t - \hat{P}^{(i-1)}_{t} \|_2^2 \mid X_\tau, y^*_{<t} \right].
\end{aligned}
\end{equation}
\end{corollary}

\subsection{Setup and Preliminary Lemmas}

We focus on the training of a single successor model $m_i$ with parameters $\theta_i \in \mathbb{R}^d$ at an arbitrary decoding step $t$.

\begin{itemize}
    \item \textbf{Losses:} The total loss for token $t$ is $\mathcal{L}_{i,t}(\theta_i) = \mathcal{L}_{s,t}(\theta_i) + \alpha \mathcal{L}_{ce,t}(\theta_i)$, where $\mathcal{L}_{s,t}$ is the error-suppression loss and $\mathcal{L}_{ce,t}(\theta_i) = - \log p^{(i)}_t(y^*_t; \theta_i)$ is the cross-entropy loss.
    \item \textbf{Gradients:} Let $g_{s,t}(\theta_i) = \nabla_{\theta_i} \mathcal{L}_{s,t}(\theta_i)$, $g_{ce,t}(\theta_i) = \nabla_{\theta_i} \mathcal{L}_{ce,t}(\theta_i)$, and the total gradient is $g_{total,t}(\theta_i) = g_{s,t}(\theta_i) + \alpha g_{ce,t}(\theta_i)$.
    \item \textbf{Logits and Jacobian:} Let $z^{(i)}_t$ be the logits from model $m_i$. Let $J_i(\theta_i) = \frac{\partial z^{(i)}_t(\theta_i)}{\partial \theta_i}$ be the Jacobian of the logit function.
\end{itemize}

We begin by characterizing the exact gradients of the loss components.

\begin{lemma}[Loss Gradients in Logit Space]
\label{lemma:logit_grads}
At a given step $t$, let $p^{(i)}_t = \mathrm{softmax}(z^{(i)}_t)$. Let $y^*_t$ be the one-hot vector for the correct token, and if the predecessor erred, let $y_{err}$ be the one-hot vector for the $\mathrm{ErrorToken}^{(i-1)}_t$. The gradients of the loss components with respect to the logits $z^{(i)}_t$ are:
\begin{align}
\nabla_{z^{(i)}_t} \mathcal{L}_{ce,t} &= p^{(i)}_t - y^*_t, \\
\nabla_{z^{(i)}_t} \mathcal{L}_{s,t} &= \beta \sigma(-u(\theta_i)) (y_{err} - y^*_t) \cdot \mathbf{1}\{ \mathrm{ErrorToken}^{(i-1)}_t \neq \emptyset \},
\end{align}
where $u(\theta_i) = \beta[\log p^{(i)}_t(y^*_t) - \log p^{(i)}_t(\mathrm{ErrorToken}^{(i-1)}_t)]$.
\end{lemma}
\begin{proof}
The gradient for $\mathcal{L}_{ce,t}$ is a standard result. For $\mathcal{L}_{s,t}$, we apply the chain rule $\nabla_z \mathcal{L}_{s,t} = \frac{\partial \mathcal{L}_{s,t}}{\partial u} \frac{\partial u}{\partial z}$. We have $\frac{\partial \mathcal{L}_{s,t}}{\partial u} = -\sigma(-u)$. For the second term:
\begin{equation}
\frac{\partial u}{\partial z^{(i)}_t} = \beta \left( \frac{\partial \log p^{(i)}_t(y^*_t)}{\partial z^{(i)}_t} - \frac{\partial \log p^{(i)}_t(y_{err})}{\partial z^{(i)}_t} \right).
\end{equation}

Using the identity $\frac{\partial \log p_k}{\partial z} = \mathbf{e}_k - p$, where $\mathbf{e}_k$ is the one-hot vector for class $k$:
\begin{equation}
\frac{\partial u}{\partial z^{(i)}_t} = \beta \left( (y^*_t - p^{(i)}_t) - (y_{err} - p^{(i)}_t) \right) = \beta (y^*_t - y_{err}).
\end{equation}
Combining these gives $\nabla_{z^{(i)}_t} \mathcal{L}_{s,t} = -\sigma(-u) \cdot \beta (y^*_t - y_{err}) = \beta \sigma(-u) (y_{err} - y^*_t)$, which proves the result for the case when the predecessor errs. Otherwise, the gradient is zero.
\end{proof}
\begin{corollary}[Loss Gradients in Parameter Space]
The gradients with respect to the parameters $\theta_i$ are obtained via the Jacobian:
$g_{ce,t}(\theta_i) = J_i(\theta_i)^\top (p^{(i)}_t - y^*_t)$ and $g_{s,t}(\theta_i) = J_i(\theta_i)^\top \left[ \beta \sigma(-u) (y_{err} - y^*_t) \cdot \mathbf{1}\{\mathrm{ErrorToken}^{(i-1)}_t \neq \emptyset\} \right]$.
\end{corollary}

\subsection{Gradient Alignment and Main Result}

To formalize the argument that our training objective provides a valid descent path, we introduce two standard assumptions regarding the optimization landscape. We analyze the gradients of the primary cross-entropy loss, $g_{ce,t}(\theta_i)$, and the error-suppression loss, $g_{s,t}(\theta_i)$. The total gradient used for the update is $g_{total,t}(\theta_i) = \alpha g_{ce,t}(\theta_i) + g_{s,t}(\theta_i)$.

\begin{assumption}[Smoothness and Parameter Compactness]
\label{ass:smoothness}
The primary loss function, $\mathcal{L}_{ce,t}(\theta_i)$, is $L$-smooth for some constant $L>0$, meaning its gradient is Lipschitz continuous with constant $L$. Furthermore, we assume the training trajectory of the parameters $\theta_i$ remains within a compact set.
\end{assumption}

\begin{assumption}[Gradient Relative Alignment and Bounded Norm]
\label{ass:grad_align}
During the training of successor $m_i$, for the subset of training data where the predecessor model $m_{i-1}$ predicts incorrectly, we assume there exist constants $\rho_i \in [0, 1)$ and $\Gamma_i > 0$ that characterize the geometric relationship between the two gradient components for all parameters $\theta_i$ encountered during training:
\begin{align}
\label{eq:grad_align_angle}
\langle g_{s,t}(\theta_i), g_{ce,t}(\theta_i) \rangle &\ge -\rho_i \|g_{s,t}(\theta_i)\| \|g_{ce,t}(\theta_i)\|, \\
\label{eq:grad_align_norm}
\|g_{s,t}(\theta_i)\| &\le \Gamma_i \|g_{ce,t}(\theta_i)\|.
\end{align}
\end{assumption}
\noindent\textbf{Remark.} The first condition, relative alignment, bounds the degree to which the error-suppression loss can oppose the primary training signal. A value of $\rho_i$ close to 0 indicates strong alignment. The second condition bounds the relative magnitude of the two gradients. Together, they formalize the notion that $\mathcal{L}_{s,t}$ acts as a helpful, non-contradictory regularizer.

\begin{theorem}[Guaranteed Descent on Primary Loss]
\label{thm:descent}
Under Assumptions \ref{ass:smoothness} and \ref{ass:grad_align}, there exists a constructive range for the hyperparameter $\alpha$ and the learning rate $\eta$ that guarantees a decrease in the primary cross-entropy loss $\mathcal{L}_{ce,t}$ with each gradient descent step $\theta'_i = \theta_i - \eta g_{total,t}(\theta_i)$, provided $g_{ce,t}(\theta_i) \neq 0$.
\end{theorem}
\begin{proof}
The $L$-smoothness of $\mathcal{L}_{ce,t}$ provides the following descent lemma:
\begin{equation}
\mathcal{L}_{ce,t}(\theta'_i) \le \mathcal{L}_{ce,t}(\theta_i) - \eta \langle g_{total,t}(\theta_i), g_{ce,t}(\theta_i) \rangle + \frac{L\eta^2}{2} \|g_{total,t}(\theta_i)\|^2.
\end{equation}
To guarantee a decrease in $\mathcal{L}_{ce,t}$, a sufficiently small learning rate $\eta$ can be chosen provided that the inner product $\langle g_{total,t}, g_{ce,t} \rangle$ is strictly positive. We analyze this inner product based on the predecessor's prediction.

\textbf{Case 1: Predecessor is correct ($\mathrm{ErrorToken}^{(i-1)}_t = \emptyset$).}
In this case, $g_{s,t}(\theta_i) = 0$. The total gradient simplifies to $g_{total,t}(\theta_i) = \alpha g_{ce,t}(\theta_i)$. The inner product becomes:
\begin{equation}
\langle g_{total,t}, g_{ce,t} \rangle = \langle \alpha g_{ce,t}, g_{ce,t} \rangle = \alpha \|g_{ce,t}\|^2.
\end{equation}
Since we assume $g_{ce,t} \neq 0$ and $\alpha > 0$, this inner product is strictly positive.

\textbf{Case 2: Predecessor is incorrect ($\mathrm{ErrorToken}^{(i-1)}_t \neq \emptyset$).}
The full gradient is active. We must find conditions under which the full inner product is positive.

\textbf{Step 1: Lower bound the inner product $\langle g_{total,t}, g_{ce,t} \rangle$.}
\begin{align}
\langle g_{total,t}, g_{ce,t} \rangle &= \langle \alpha g_{ce,t} + g_{s,t}, g_{ce,t} \rangle = \alpha \|g_{ce,t}\|^2 + \langle g_{s,t}, g_{ce,t} \rangle \\
&\ge \alpha \|g_{ce,t}\|^2 - \rho_i \|g_{s,t}\| \|g_{ce,t}\| \quad (\text{by Eq. } \ref{eq:grad_align_angle}) \\
&\ge \alpha \|g_{ce,t}\|^2 - \rho_i (\Gamma_i \|g_{ce,t}\|) \|g_{ce,t}\| \quad (\text{by Eq. } \ref{eq:grad_align_norm}) \\
&= (\alpha - \rho_i \Gamma_i) \|g_{ce,t}\|^2.
\end{align}
For this inner product to be strictly positive, we require $\alpha - \rho_i \Gamma_i > 0$. This gives us a lower bound for our hyperparameter $\alpha$:
$$ \alpha > \rho_i \Gamma_i. $$

\textbf{Step 2: Upper bound the total gradient norm $\|g_{total,t}\|^2$.}
\begin{align}
\|g_{total,t}\|^2 &= \|\alpha g_{ce,t} + g_{s,t}\|^2 \le (\|\alpha g_{ce,t}\| + \|g_{s,t}\|)^2 \\
&\le (\alpha \|g_{ce,t}\| + \Gamma_i \|g_{ce,t}\|)^2 \quad (\text{by Eq. } \ref{eq:grad_align_norm}) \\
&= (\alpha + \Gamma_i)^2 \|g_{ce,t}\|^2.
\end{align}

\textbf{Step 3: Construct the learning rate bound.}
With a choice of $\alpha > \rho_i \Gamma_i$, the inner product $\langle g_{total,t}, g_{ce,t} \rangle$ is guaranteed to be positive. To ensure descent, the learning rate $\eta$ must be chosen in the interval:
\begin{equation}
0 < \eta < \frac{2 \langle g_{total,t}, g_{ce,t} \rangle}{L \|g_{total,t}\|^2}.
\end{equation}
To find a constructive, safe range for $\eta$, we can establish a lower bound for the right-hand side of the inequality. Using the lower bound for the numerator (from Step 1) and the upper bound for the denominator's norm term (from Step 2), we have:
\begin{equation}
\frac{2 \langle g_{total,t}, g_{ce,t} \rangle}{L \|g_{total,t}\|^2} \ge \frac{2 (\alpha - \rho_i \Gamma_i) \|g_{ce,t}\|^2}{L (\alpha + \Gamma_i)^2 \|g_{ce,t}\|^2} = \frac{2 (\alpha - \rho_i \Gamma_i)}{L (\alpha + \Gamma_i)^2}.
\end{equation}

Let us define this constructive lower bound as $\eta^*(\alpha) = \frac{2 (\alpha - \rho_i \Gamma_i)}{L (\alpha + \Gamma_i)^2}$. For any $\alpha > \rho_i \Gamma_i$, the numerator is positive, so $\eta^*(\alpha) > 0$. 
Therefore, by choosing any learning rate $\eta$ such that $0 < \eta < \eta^*(\alpha)$, we guarantee that it is within the valid range for descent, ensuring that $\mathcal{L}_{ce,t}(\theta'_i) < \mathcal{L}_{ce,t}(\theta_i)$.

By combining both cases, we have shown that a valid range for $\alpha$ and $\eta$ exists to ensure descent on the primary cross-entropy loss.
\end{proof}

\subsection{Connecting the Optimization View to the Core Assumption}
The results from this optimization analysis provide a strong heuristic basis for Assumption \ref{ass:correction_efficacy_training}. Theorem \ref{thm:descent} demonstrates that our composite loss function, under the geometric conditions of Assumption \ref{ass:grad_align}, is designed to effectively minimize the primary cross-entropy loss $\mathcal{L}_{ce,t}$. A reduction in $\mathcal{L}_{ce,t}$ implies that the model's prediction $p^{(i)}_t$ moves closer to the ground truth $y^*_t$. This is the fundamental mechanism that drives the model $m_i$ to become an effective corrector of the predecessor's residual error, thereby satisfying the conditions required for the MSE reduction theorems.

\end{document}